%% file: main.tex
\documentclass{article}

\usepackage[main,final]{neurips_2026}
\usepackage{comment}
\usepackage{wrapfig}
\usepackage[utf8]{inputenc} 
\usepackage[T1]{fontenc}    
\usepackage{hyperref}       
\usepackage{url}            
\usepackage{booktabs}       
\usepackage{amsfonts}       
\usepackage{nicefrac}       
\usepackage{microtype}      
\usepackage{xcolor}         
\usepackage{pifont}
\usepackage{multirow}
\usepackage{bbm}
\usepackage{makecell}
\include{math_commands}

\usepackage{graphicx}
\usepackage{subcaption}
\usepackage{booktabs}

\usepackage{amsmath}
\usepackage{amssymb}
\usepackage{mathtools}
\usepackage{amsthm}
\RequirePackage{algorithm}
\RequirePackage{algorithmic}

\usepackage[capitalize,noabbrev]{cleveref}
\usepackage{physics}

\theoremstyle{plain}
\newtheorem{theorem}{Theorem}[section]
\newtheorem{proposition}[theorem]{Proposition}

\theoremstyle{definition}

\theoremstyle{remark}

\usepackage[disable,textsize=tiny]{todonotes}

\newcommand{\Ehull}{E_{\text{hull}}}
\definecolor{isblue}{RGB}{0,90,181}
\definecolor{mored}{RGB}{178,24,43}
\definecolor{cmtgreen}{RGB}{96,128,96}
\newcommand{\tIS}{\hfill{\footnotesize\textcolor{isblue}{[IS]}}}
\newcommand{\tMO}{\hfill{\footnotesize\textcolor{mored}{[MO]}}}
\newcommand{\tISMO}{\hfill{\footnotesize\textcolor{isblue}{[IS]}\,\textcolor{mored}{[MO]}}}
\newcommand{\is}[1]{\textcolor{isblue}{#1}}
\newcommand{\mo}[1]{\textcolor{mored}{#1}}
\newcommand{\cmt}[1]{\STATE \textcolor{cmtgreen}{\textit{// #1}}}
\newcommand{\greencheck}{\textcolor{green!50!black}{\ding{51}}}
\newcommand{\redx}{\textcolor{red}{\ding{55}}}
\title{Diffusion Tree Sampling for Inference Time Adaptation of Crystal Generative Models}

\author{%
  Daniel T. Levy$^{*}$ $\quad$ Vineet Jain\thanks{Equal contribution. Correspondence: daniel.levy6@mail.mcgill.ca. Code is available at: \url{https://github.com/danielTLevy/MaterialDTS}} $\quad$ Tara Akhound-Sadegh \\ \textbf{Sékou-Oumar Kaba} $\quad$ \textbf{Siamak Ravanbakhsh} \vspace{0.5em}\\ 
    McGill University \\
    Mila - Quebec Artificial Intelligence Institute
}

\begin{document}

\maketitle

\begin{abstract}
Diffusion-based generative models have become strong priors over plausible crystal structures, but turning that prior into promising candidates requires steering toward several conflicting properties at once.
Fine-tuning is expensive and produces a new model for every new objective, while differentiable guidance is not applicable to many real-world tasks. We instead adapt these generative models at \emph{inference time} using Diffusion Tree Sampling (DTS), a Monte Carlo tree search procedure over the denoising trajectory. We make two technical contributions on top of DTS that are particularly relevant for AI-for-science applications. First, we show how to use existing property-conditional generative models (e.g., models fine-tuned for classifier-free guidance) as guided proposals inside DTS, without distorting the target distribution. Second, we extend DTS to multi-objective sampling, so that a single search produces samples spanning an entire Pareto front. We instantiate both adaptations on de novo crystal generation: improving the stability of crystals generated by different models, using pretrained conditional models to generate crystals belonging to rare space groups or with high bulk modulus, and generating crystals along a Pareto front of high dielectric constant and high bandgap materials. Across these tasks, DTS-based inference-time adaptation successfully improves model outputs without modifying the models themselves.
\end{abstract}

\section{Introduction}

Foundation models are flourishing in AI for science due to the superior generalization of large models trained on massive datasets \citep{kaplan2020scaling,bommasani2021opportunities}. In materials discovery, this has motivated powerful diffusion-based generative models that capture the distribution of plausible crystal structures: MatterGen \citep{zeni2023mattergen} and DiffCSP \citep{jiao2023crystal} jointly model atom types, lattice, and coordinates, and can sample chemically plausible crystals. Yet, despite strong sample quality, these models have rarely translated into confirmed new discoveries.

A key barrier is that unconditional generation over all possible crystals is generally not directly useful. Candidate materials should not just be sampled from the manifold of all materials, but rather obtained through exploration of structures that are stable, sit in a particular property window, and trade off several conflicting properties (functionality, cost, element availability). Existing strategies for diffusion models typically modify the base model: fine-tuning property-conditional variants \citep{zeni2023mattergen}, reward-conditioned RL \citep{chen2025matinvent}, or active learning loops \citep{li2025active}. Each requires labeled data, training compute, and produces a new model per design problem, which is at odds with the foundation-model paradigm.

Inference-time adaptation offers a different path: leave the pretrained model fixed, and steer its sampling toward the user's specific goals \citep{dhariwal2021diffusion,ho2022classifier,yeh2024training,tang2024inference,jain2025diffusion,uehara2025inference}. This is well suited to materials, where the validation of candidate stability with density functional theory (DFT) calculations and the subsequent attempt to synthesize them is orders of magnitude more expensive than generation. Therefore, allocating inference compute to candidate quality is far more valuable than producing high volumes of low-quality structures. Among inference-time methods, Diffusion Tree Sampling (DTS) \citep{jain2025diffusion} casts denoising as a tree of partial trajectories, allocating compute adaptively to promising branches via Monte Carlo Tree Search (MCTS).

Two features of materials candidate generation are not addressed by the original DTS formulation, and motivate this work. First, in many cases, there \emph{are} property-conditional models -- e.g., MatterGen has variants for individual properties -- which are dramatically more effective than the unconditional model at finding samples with a given property. Naively swapping in a conditional sampler inside DTS tilts towards the reward twice, resulting in a different distribution from the target. Second, real design problems are rarely single-objective: properties trade off, and the appropriate target is the \emph{Pareto front} of the conflicting objectives \citep{miettinen1999nonlinear,deb2016multi,riebesell2024pushing}.

\begin{figure}
    \centering
    \includegraphics[width=\linewidth]{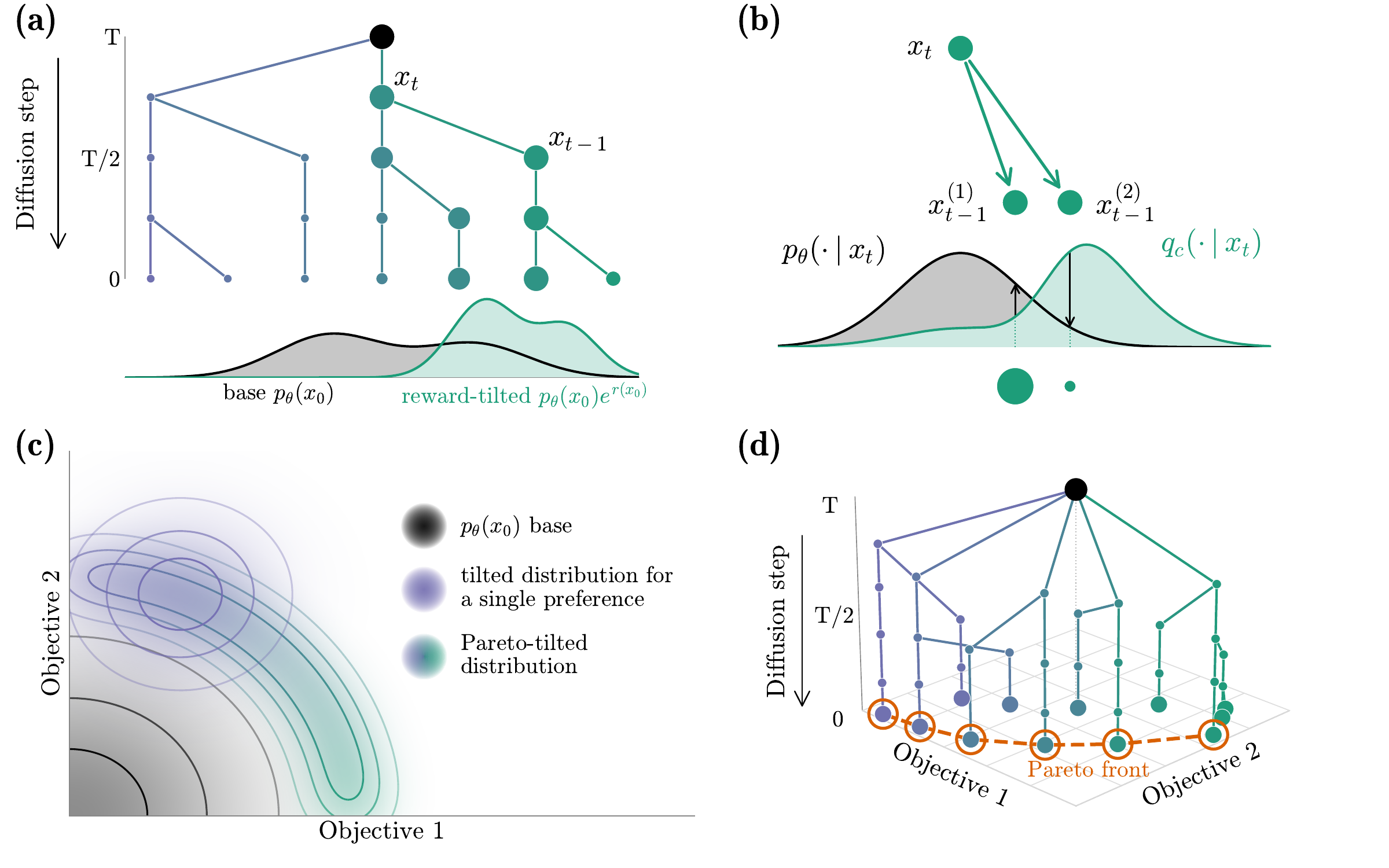}
    \caption{Overview of the method. \textbf{(a)}: DTS treats intermediate states of denoising diffusion as nodes in a tree, and steers sampling towards a reward-tilted distribution. Each node is attached to an estimate of expected reward (denoted here by node size). \textbf{(b)}: We may use a different model $q_c$ that gives better denoising proposals, but we need to correct its influence by reweighting nodes by the proposal's ratio with the base model. \textbf{(c)}: In the multi-objective setting, we wish to sample from a distribution obtained by integrating reward-tilted distributions over different preferences between objectives. \textbf{(d)}: We use a single DTS tree to sample from this multi-objective target, conditioning each pass through the tree on a different preference.}
    \label{fig:champion}
    \vspace{-1em}
\end{figure}

\paragraph{Contributions.}
\textbf{(1)} \emph{Conditional generative models inside DTS.} We show how to use property-conditional diffusion models (e.g. fine-tuned variants of MatterGen) as guided proposals within DTS, while keeping the reward-tilted unconditional base model as the target. This uses an importance-corrected backup and a corrected selection rule.
\textbf{(2)} \emph{Multi-objective DTS.} We formalize multi-objective sampling from a diffusion model as sampling from a mixture of reward-tilted distributions, each corresponding to a different preference over objectives.  We extend DTS to this setting by maintaining per-objective soft value estimates and selecting expansions through sampled preferences, so that a single search samples the Pareto front while sharing computation across preferences and across different conditioning.
\textbf{(3)} \emph{Application to materials discovery.}
We use the method to improve de novo crystal generation across multiple tasks: we show improved results in generating stable crystals, we demonstrate the ability to generate crystals with specific features using pretrained conditional models, and we demonstrate how to sample crystals along a Pareto front of conflicting objectives, all without having to modify any model weights or use reward gradients. 

\section{Background}

\subsection{Diffusion models for materials discovery}
Conditional materials generation aims to propose previously unknown crystalline compounds with desirable physical properties. The space of possible structures is combinatorially large, spanning composition, stoichiometry, lattice geometry, and atomic coordinates; only a small fraction is thermodynamically stable, and a smaller fraction still exhibits any given target property. We focus on \emph{de novo generation}, which jointly proposes composition, lattice, and coordinates. Stability of candidates can be validated computationally by DFT, which is orders of magnitude more expensive than generation. Classical pipelines combine heuristic generation with first-principles relaxation \citep{oganov2006crystal,pickard2011ab}. Some methods search using basin hopping \citep{wales1997global,wales1999global} -- a structure with a natural analogue in the tree-search method we use (see Appendix \ref{app:basin}).

A recent line of work trains generative models directly on databases of known crystals; diffusion models are a strong fit since they jointly model the discrete (atom types) and continuous (lattice and coordinates) components of a crystal. We focus on two models. \textbf{DiffCSP} \citep{jiao2023crystal} parameterizes a joint diffusion process over fractional coordinates, lattice, and atom types, and supports time-conditional classifier-based guidance. \textbf{MatterGen} \citep{zeni2023mattergen} is trained on a substantially larger dataset and provides adapter modules for efficient fine-tuning into property-conditional models, sampled with classifier-free guidance \citep{ho2022classifier}. Both suffer from a high rate of false positives: only a small fraction of generated structures are predicted to be stable, and a smaller fraction yet satisfies property constraints. Existing strategies modify the base model via classifier-based guidance \citep{jiao2023crystal}, fine-tuned conditional variants \citep{zeni2023mattergen}, RL \citep{chen2025matinvent}, and active learning \citep{li2025active}. Our work is complementary: it steers a fixed pretrained generative model at inference time, and extends this steering to settings where conditional variants exist and where multiple objectives must be balanced.

\subsection{Inference-time steering and Diffusion Tree Sampling}
Let $p_\theta$ denote a pretrained diffusion model with reverse transitions $p_\theta(\rvx_{t-1}\mid \rvx_t)$, inducing a prior $\rvx_0 \sim p_\theta(\rvx_0)$. Inference-time steering introduces a reward $r(\rvx_0)$ available only after full denoising, and aims to sample from the \emph{reward-tilted} target $\pi^\star(\rvx_0) = \tfrac{1}{Z} p_\theta(\rvx_0)\exp(\lambda r(\rvx_0))$, which is the KL-regularized optimum of expected reward relative to the base model with inverse temperature $\lambda$. A useful characterization of expected reward at a noisy state $\rvx_t$ is the \emph{soft value function} 
$v_t(\rvx_t) = \tfrac{1}{\lambda}\log \E[\exp(\lambda r(\rvx_0))\mid \rvx_t]$ with $v_0(\rvx_0) = r(\rvx_0)$, which satisfies the soft Bellman recursion
\begin{equation}\label{eq:value}
v_t(\rvx_t)\;=\;\frac{1}{\lambda}\log \E_{p_\theta(\rvx_{t-1}\mid \rvx_t)}\!\left[\exp(\lambda v_{t-1}(\rvx_{t-1}))\right].
\end{equation}
The optimal aligned reverse transition is $\pi^\star_t(\rvx_{t-1}\mid \rvx_t)\propto p_\theta(\rvx_{t-1}\mid \rvx_t)\exp(\lambda v_{t-1}(\rvx_{t-1}))$. Most inference-time steering methods approximate $v_{t-1}$ via gradient guidance, SMC, or local search over denoising proposals \citep{uehara2025inference}.

A central difficulty is \emph{value estimation at high noise}: common approximations become unreliable early in the reverse process. \emph{Diffusion Tree Sampling} (DTS) \citep{jain2025diffusion} addresses this by casting denoising as a finite-horizon tree whose nodes are states $\rvx_t$ with value estimates $\hat  v_t(\rvx_t)$ and whose edges are reverse transitions. The algorithm repeatedly builds a tree by starting at a root node $\rvx_{T+1}$ and performing (1) \emph{selection} of children $\propto \exp(\lambda \hat v)$ until reaching a node that should expand, (2) \emph{expansion} by sampling a new child from $p_\theta$, (3) \emph{rollout} of the new child to a terminal $\rvx_0$, assigning $\hat v(\rvx_0) = r(\rvx_0)$ and (4) \emph{backup} via the soft Bellman log-sum-exp. Reusing these backed-up values yields an anytime procedure whose empirical leaf distribution converges to $\pi^\star$; a greedy variant replaces sampling with argmax selection.

\section{Adapting Diffusion Tree Sampling to Crystal Generative Models}

Standard DTS assumes (i) a single scalar reward, and (ii) that the reverse transitions used during expansion and rollout are exactly the unconditional base model. Both assumptions are limiting in the materials setting. In this section we adapt DTS to relax them: \cref{sec:cond} shows how to use existing property-conditional generative models as guided proposals inside DTS without changing the target distribution; \cref{sec:modts} extends the search to multi-objective sampling. The full algorithm is given in Appendix \ref{app:algo}.

\subsection{Adaptation 1: leveraging conditional models inside DTS}
\label{sec:cond}

When available, a \emph{property-guided} diffusion model -- e.g., a conditional model trained for classifier-free guidance (CFG) -- can be much more effective at \emph{locally} steering denoising steps toward narrow high-reward regions than a purely reward-based search on top of the unconditional model.
For example, MatterGen provides property-conditional variants for individual properties (bandgap, space group, etc.), and we would like to use them as proposals inside DTS without losing the well-defined target distribution induced by the unconditional base model.

We denote the unconditional reverse transition by $p_\theta(\rvx_{t-1}\mid \rvx_t)$ and a guided/conditional transition by $p_\theta(\rvx_{t-1}\mid \rvx_t, c)$, where $c$ encodes a target (e.g. a desired property). 
In our setting, we assume that providing conditioning information $c$ helps with optimizing some scalar reward $r(\cdot)$ (e.g. conditioning on a target property).

\paragraph{The double-tilt problem.}
A naive attempt would be to replace the base model inside DTS with the conditional model and then tilt by $\exp(\lambda r(\rvx_0))$. This would target $\pi_{\text{naive}}(\rvx_0)\propto p_\theta(\rvx_0\mid c)\,\exp(\lambda r(\rvx_0))$, which would tilt twice: the conditional model already biases mass toward the properties encoded by $c$, and the additional Boltzmann tilt amplifies the same preference, yielding a different target than the original $\pi^\star$. An example of this issue is shown in Appendix \ref{app:toy}, using a Gaussian-mixture toy experiment.

\paragraph{Importance-corrected backup.}
Our solution is to keep the \emph{unconditional} model as the target and use the conditional model only as a proposal, correcting via importance sampling (IS). Concretely, during expansion and rollout from $\rvx_t$ we sample $\rvx_{t-1}$ from the guided proposal $q_c(\rvx_{t-1}\mid\rvx_t) \coloneqq p_\theta(\rvx_{t-1}\mid\rvx_t, c)$ instead of $p_\theta(\rvx_{t-1}\mid\rvx_t)$, and attach to the new node the \emph{importance weight}
\begin{equation}
\label{eq:eta}
\eta(\rvx_{t-1}) \coloneqq \frac{p_\theta(\rvx_{t-1}\mid\rvx_t)}{q_c(\rvx_{t-1}\mid\rvx_t)},
\end{equation}
the likelihood ratio of base to proposal. Models trained for CFG are convenient here because they provide both transitions. The soft value of \cref{eq:value} is then an expectation under $q_c$ weighted by $\eta$,
\begin{equation}
\label{eq:is_backup}
v_t(\rvx_t) = \frac{1}{\lambda}\log \E_{\rvx_{t-1}\sim q_c(\cdot\mid\rvx_t)}\!\big[\eta(\rvx_{t-1})\exp\!\big(\lambda v_{t-1}(\rvx_{t-1})\big)\big],
\end{equation}
so backups remain consistent with the unconditional base dynamics.

\paragraph{Importance correction in selection.}
Selection among existing children must be corrected for the same reason, since the children were drawn from $q_c$. We therefore select a child $\rvz\in\mathcal C(\rvx_t)$ with probability
\begin{equation}
\label{eq:select_is}
\propto\; \eta(\rvz)\,\exp\!\big(\lambda\,\hat v(\rvz)\big),
\end{equation}
which is self-normalized importance sampling of the aligned base transition $p_\theta(\rvx_{t-1}\mid\rvx_t)\exp(\lambda v_{t-1}(\rvx_{t-1}))$. Without $\eta$, selection would be biased toward $q_c$ and the tree would converge to a target that depends on the proposal.

\subsection{Adaptation 2: Multi-Objective Diffusion Tree Sampling (MO-DTS)}
\label{sec:modts}
Many materials problems involve \emph{multiple} competing goals. Let $\rvr(\rvx) \coloneqq  [r_1(\rvx),\dots,r_m(\rvx)]$ be a vector of $m$ objectives to be maximized; a point $\rvx^\star$ \emph{dominates} $\rvx$ if it is at least as good on all objectives and strictly better on one, and the set of non-dominated points is the \emph{Pareto front}. Multi-objective optimization can be tackled by taking a \emph{preference} over objectives $w \in \Delta^{m-1}$ that encodes a trade-off, and a \emph{scalarization}, which is a map $s:\sR^m \cross \Delta^{m-1}\to\sR$ that turns a reward vector into a scalar according to $w$.
The simplest choice is linear scalarization $s(\rva \mid w)\coloneqq w^{\top}\rva$, although other choices are possible (see Appendix \ref{app:scalarization}).
For a fixed $w$ this defines the scalar reward $r(\rvx \mid w)\coloneqq s(\rvr(\rvx) \mid w)$, and the multi-objective problem is reduced to a set of single-objective problems: $\max_{\rvx} r(\rvx \mid w)$.

\paragraph{Sampling from the Pareto front.}
Given objectives $\rvr(\rvx) \in \sR^{m}$ and a scalarized $r(\rvx\mid w)$, we define the target of \emph{sampling from the Pareto front} as first sampling a preference $w\sim p(w)$ where $p(w)$ is a user-chosen distribution over preferences (e.g., Dirichlet on the simplex), then sampling
$\rvx_0$ from the corresponding reward-tilted base distribution:
\[
\pi^\star(\rvx_0\mid w)
=
p_\theta(\rvx_0)\exp(\lambda r(\rvx_0\mid w))/Z(w),
\qquad
Z(w)=\E_{\rvx_0\sim p_\theta}
\left[\exp(\lambda r(\rvx_0\mid w))\right].
\]
Marginalizing out $w$ yields a mixture of reward-tilted distributions,
\begin{equation}
\pi^\star(\rvx_0)
=
p_\theta(\rvx_0)\,
\E_{w\sim p(w)}
\left[\exp(\lambda r(\rvx_0\mid w))/Z(w)\right],
\label{eq:pareto_mixture}
\end{equation}
which assigns high probability to candidates that are strong under \emph{some} trade-off, while remaining plausible under the base model. 

\paragraph{A single shared tree.}
When lifting a single-objective method to the multi-objective setting, treating each scalarized objective $r(\cdot \mid w)$ in isolation is highly inefficient: a key design requirement is to share computation across preferences. It is non-trivial to adapt arbitrary alignment methods (Best-of-$N$, SMC \citep{singhal2025general}) under this requirement. DTS \citep{jain2025diffusion} can be extended rather directly to approximately sample from \cref{eq:pareto_mixture} while sharing computation: maintain a separate value estimate per objective during backup, and use a \emph{scalarized value} based on these estimates during selection.

Each terminal rollout produces a vector reward $\rvr(\rvx_0)$, and we maintain at each node a value vector $\rvv_t(\rvx_t) = [v_{t,1}(\rvx_t),\dots,v_{t,m}(\rvx_t)]$, with each $v_{t,i}$ updated via the soft backup of \cref{eq:value} applied to objective $i$. Each tree traversal first samples $w \sim p(w)$, then computes a scalarized value by applying the scalarization to the node's value vector, $v_t(\rvx_t | w) \coloneqq s(\rvv(\rvx_t) \mid w)$.

\paragraph{MO-DTS and conditioning.}
When available, we can use a conditional diffusion model where the condition is a function of the preference.
That is, for each preference $w$, our diffusion model samples from  $q_w(\rvx_{t-1}\mid \rvx_t) := p_\theta(\rvx_{t-1}\mid \rvx_t, c(w))$ where $c(w)$ is a mapping from preferences to conditions that would help maximize $r(\cdot | w)$.
In practice, this can mean conditioning the diffusion model on the features of samples in the training set that maximize $r(\cdot | w)$, as they represent reasonable targets.

At each pass through the tree, $w$ can be sampled anew. The sampled $w$ not only affects the choice of children while moving down the tree from the root, but also affects how new leaf nodes are generated. However, due to the importance correction of \cref{sec:cond}, the target distribution does not depend on the choice of $c(w)$, and values $\rvv_t(\rvx_t)$ do not depend on $w$, so the whole tree remains consistent.

\paragraph{Per-objective values as an upper-confidence approximation.}
Scalarizing at selection time is an approximation: the exact soft value of the scalarized reward, $v_t^\star(\rvx_t\mid w) \coloneqq \tfrac1\lambda\log\E[\exp(\lambda\,r(\rvx_0\mid w))\mid\rvx_t]$, would require a separate backup for every $w$, whereas MO-DTS backs up each objective once and uses $v_t(\rvx_t\mid w)=s(\rvv_t(\rvx_t)\mid w)$. In Appendix \ref{app:approx} we show that the surrogate is optimistic: it upper-bounds $v_t^\star$ for linear scalarizations at every $\lambda$, and coincides with $v_t^\star$ at $t=0$. The gap is governed by how far the objectives are from co-varying under the denoising distribution, so selection is most optimistic about nodes whose objectives trade off; MO-DTS therefore provides a search heuristic for \cref{eq:pareto_mixture} rather than an exact sampler for every $w$.

\subsection{Coarse branching and parallelization}
\label{sec:parallel}
Following \citet{jain2025diffusion}, we restrict branching to a set of \emph{branching times} $\mathcal T\subseteq\{1,\dots,T+1\}$. A node $\rvx_t$ is \emph{expandable} if $t\in\mathcal T$ and progressive widening permits, $|\mathcal C(\rvx_t)|<C\,N(\rvx_t)^{\alpha}$, where $\mathcal{C}$ is the children of a node, $N$ is the number of visits so far, and $C$ and $\alpha$ are hyperparameters. Nodes at other times are created only during rollouts and have a single child, so selection passes through them deterministically. In practice, we set the branching times to be regularly spaced every $L$ timesteps. 
At a matched number of function evaluations (NFEs), DTS produces more samples due to trajectory reuse.
The search also admits a batched implementation: $B$ traversals advance synchronously from the root node, and all rollout steps at the same $t$ are batched into one call to the sampler, so each pass through the tree generates $B$ leaf nodes. Throughput is thus at least as high as generating $B$ independent trajectories as in best-of-$N$.
\section{Application to Materials Discovery}
We instantiate the framework on four crystal-generation problems chosen to demonstrate both technical contributions across two generative models. First, we demonstrate how inference-time steering with regular DTS improves models' ability to generate stable crystals. Next, symmetry-conditioned generation targeting rare space groups (\cref{sec:symmetry}) demonstrates the benefit of using conditional models and exercises the importance-corrected guided proposals of \cref{sec:cond}. A multi-objective experiment targeting the bandgap and dielectric constant tradeoff demonstrates the utility of building a single tree across different preferences in \cref{sec:dielectric}. We examine the effect of a mismatched proposal distribution and the use of importance-correction in \cref{sec:density}, targeting high bulk modulus materials.

\subsection{De novo stability search }
\label{sec:stability}

A central challenge in de novo crystal generation is producing \emph{thermodynamically stable} structures, quantified by the \emph{energy above the convex hull} ($\Ehull$): for a given composition, the convex hull is the lower envelope of formation energies across all known competing phases. $\Ehull{=}0$ is stable; small positive $\Ehull$ (typically below $0.1$ eV/atom) is \emph{metastable}.
Current generative models produce stable samples only a small fraction of the time, bottlenecking their usefulness \citep{betala2025lemat}.
$\Ehull$ depends on the composition through the (non-differentiable) hull construction, ruling out gradient-based guidance.

We define a single-objective reward based on $\Ehull$ as predicted by MatterSim \citep{yang2024mattersim}, with the Alex-MP hull from \cite{zeni2023mattergen}. This setup reduces to DTS with scalar reward and serves as a controlled assessment of tree-search-based steering vs.\ best-of-$N$ sampling from the same generative model.
We evaluate using the common summary metric of the rate of \emph{Stable, Unique, Novel} (SUN) structures \citep{zeni2023mattergen}, where uniqueness and novelty with respect to the training set are determined using Pymatgen's \texttt{StructureMatcher} \citep{ong2013python}.

\paragraph{Results with MatterGen.}
We compare three configurations at matched NFE budgets: standard sampling from MatterGen (best-of-$N$), DTS with null reward $r(\rvx_0) = 0$, and DTS with the reward $r(\rvx_0) = -\Ehull(\rvx_0)$. The null-reward control isolates the effect of tree search itself (trajectory reuse) from the stability reward; we also vary the progressive widening parameter $\alpha$, which controls tree width. Results averaged over three seeds are shown in \cref{tab:mattergen_stable}. Tree search improves over best-of-$N$ in absolute SUN count at the same NFE budget. The energy reward consistently lowers average $\Ehull$ and improves overall stability rate. We also report the number of unique chemical systems among the resulting samples to show that diversity is retained.

\begin{table}[h]
\small
\centering
\caption{Single-objective stability search with MatterGen at matched NFE budget, averaged over three seeds. ``Null'' is DTS with $r(\rvx_0) = 0$; $\alpha$ is the progressive widening parameter.}
\resizebox{0.8\textwidth}{!}{%
\begin{tabular}{llrrrrr}
\toprule
$\alpha$ & Reward & \# Structures & \# SUN & \% Stable & SUN Unique Systems & Avg E$_{\text{hull}}$/atom \\
\midrule
\multirow{2}{*}{0.1} & Null & 5050 $\pm$ 29 & 1311 $\pm$   39 & 66.5 $\pm$ 1.1 & 912.3 $\pm$ 33.7 & 0.091 $\pm$ 0.004 \\
                              & $-\Ehull$ & 5226 $\pm$ 42 & 1432 $\pm$ 73 & 75.7 $\pm$ 0.4 & 966 $\pm$ 12 & 0.073 $\pm$ 0.000 \\
\midrule
\multirow{2}{*}{0.3} & Null & 3500 $\pm$ 0 & 1095 $\pm$ 55 & 65.3 $\pm$ 1.5 & 904.3 $\pm$ 29.4 & 0.1 $\pm$ 0.004 \\
                              & $-\Ehull$ & 3500 $\pm$ 0 & 1209 $\pm$ 31 & 70.2 $\pm$ 0.9 & 971.3 $\pm$ 29.0 & 0.085 $\pm$ 0.002 \\
\midrule
\multirow{2}{*}{0.5} & Null & 3000 $\pm$ 0 & 974 $\pm$ 27 & 66.6 $\pm$ 1.4 & 894.7 $\pm$ 27.4 & 0.090 $\pm$ 0.003 \\
                              & $-\Ehull$ & 3000 $\pm$ 0 & 1087 $\pm$ 23 & 68.6 $\pm$ 1.3 & 979.7 $\pm$ 15.2 & 0.087 $\pm$ 0.002 \\
\midrule
\multicolumn{2}{l}{MatterGen Baseline} & 2500 & 868 $\pm$ 11  & 68.4 $\pm$ 0.1 & 848.3 $\pm$ 10.9 & 0.086 $\pm$ 0.000 \\
\bottomrule
\end{tabular}
}
    \label{tab:mattergen_stable}
\end{table}

\begin{figure}[h]
    \centering
    \includegraphics[width=0.9\linewidth]{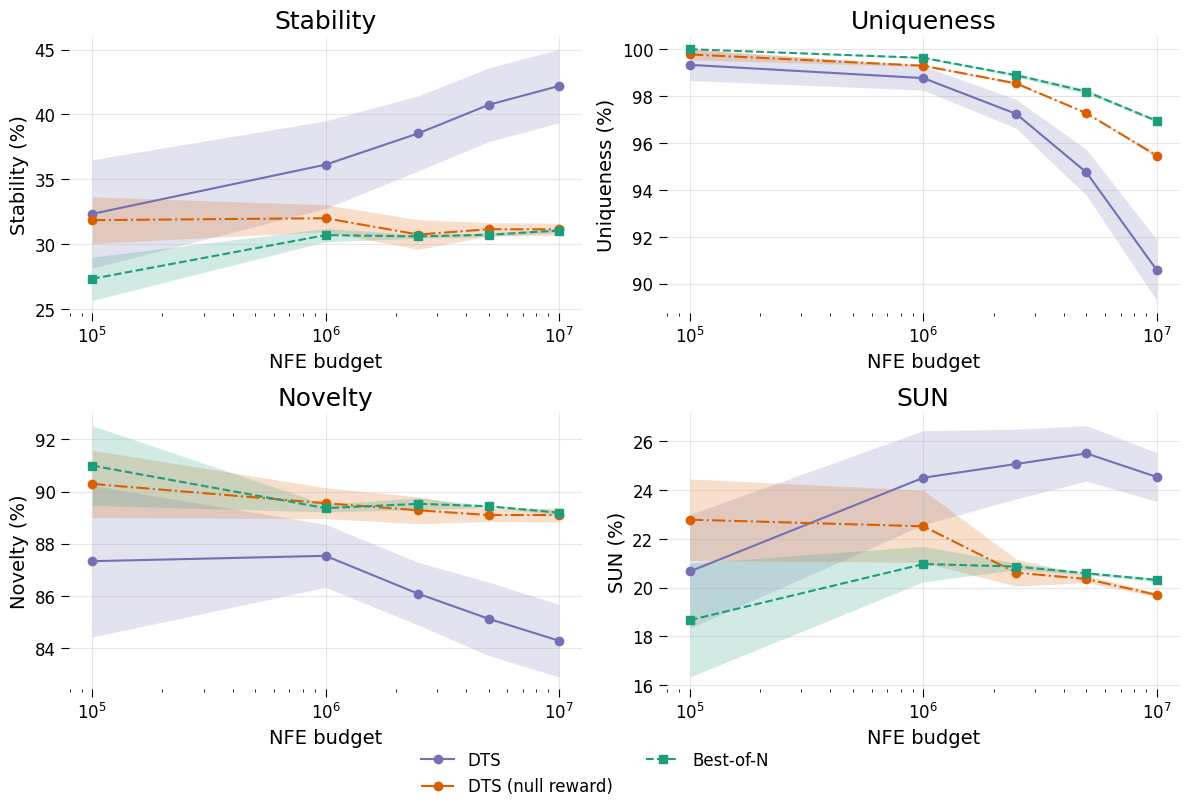}
    \caption{Inference-time scaling of DTS with DiffCSP across NFE budgets, compared with the control null-reward, and Best-of-N baselines. Standard deviations are computed across 3 seeds.}
    \label{fig:diffcspscaling}
\end{figure}

\paragraph{Results with DiffCSP and inference-time scaling.}
DiffCSP is roughly $20\times$ faster per sample than MatterGen, which lets us study how DTS scales with the NFE budget from $10^5$ to $10^7$ total NFEs (\cref{fig:diffcspscaling}). Stability rate increases with NFE budget under the $\Ehull$ reward, while the null-reward control and best-of-$N$ baselines remain roughly flat -- a clear and growing inference-time scaling benefit. Uniqueness and novelty decline slightly as compute increases, reflecting concentration of rollouts on high-reward regions.

We also compare against three inference-time steering baselines: SVDD-PM \citep{li2024derivative} (which is equivalent to TreeG-SC \citep{guo2026training} with A=1), TreeG-SD \citep{guo2026training} and rollout-based SMC \citep{singhal2025general} with two and four resampling events. All methods use the same pretrained DiffCSP model, $\Ehull$ reward, and approximately matched 1M and 10M NFE budgets;  further details are included in Appendix \ref{app:experiments_diffcsp}. The results, shown in \Cref{tab:diffcsp_matched_nfe}, demonstrate that DTS outperforms all baselines at generating stable, unique and novel crystals.

\begin{table}[t]
\centering
\caption{DiffCSP results at matched 1M and 10M NFE budgets.}
\label{tab:diffcsp_matched_nfe}
\resizebox{0.9\textwidth}{!}{%
\begin{tabular}{lrrrrrr}
\toprule
Method & Structures
& Stability (\%) $\uparrow$
& Uniqueness (\%) $\uparrow$
& Novelty (\%) $\uparrow$
& SUN (\%) $\uparrow$
& Mean $\Ehull$ $\downarrow$ \\
\midrule
\multicolumn{7}{c}{\textbf{Matched 1M NFE budget}} \\
\midrule
Best-of-$N$       & 1,000
& $30.7 \pm 0.9$ & $99.6 \pm 0.1$ & $89.4 \pm 0.3$
& $21.0 \pm 1.3$ & $0.314 \pm 0.031$ \\
SMC (2 resamples) & 554
& $30.7 \pm 0.5$ & $84.9 \pm 1.4$ & $88.8 \pm 1.0$
& $16.3 \pm 0.6$ & $0.315 \pm 0.005$ \\
SMC (4 resamples) & 332
& $29.7 \pm 3.2$ & $87.3 \pm 0.9$ & $89.3 \pm 4.5$
& $16.7 \pm 6.2$ & $0.296 \pm 0.020$ \\
SVDD-PM / TreeG-SC & 725
& $34.2 \pm 1.8$ & $99.4 \pm 0.2$ & $88.3 \pm 1.1$
& $23.2 \pm 2.2$ & $0.724 \pm 0.388$ \\
TreeG-SD         & 1,000
& $33.5 \pm 4.8$ & $99.4 \pm 0.1$ & $89.2 \pm 1.2$
& $23.2 \pm 3.4$ & $0.317 \pm 0.034$ \\
DTS (Null Reward) & 1,350
& $32.0 \pm 1.5$ & $99.3 \pm 0.1$ & $89.6 \pm 0.8$
& $22.5 \pm 2.1$ & $0.399 \pm 0.149$ \\
\textbf{DTS}              & 1,350
& $\mathbf{36.8 \pm 4.9}$ & $99.2 \pm 0.1$ & $87.5 \pm 1.4$
& $\mathbf{25.1 \pm 3.6}$ & $\mathbf{0.267 \pm 0.019}$ \\
\midrule
\multicolumn{7}{c}{\textbf{Matched 10M NFE budget}} \\
\midrule
Best-of-$N$       & 10,000
& $31.1 \pm 0.3$ & $96.9 \pm 0.1$ & $\mathbf{89.2 \pm 0.2}$
& $20.3 \pm 0.1$ & $0.435 \pm 0.035$ \\
SMC (2 resamples) & 5,549
& $31.3 \pm 0.1$ & $84.5 \pm 1.1$ & $88.6 \pm 1.4$
& $16.8 \pm 0.4$ & $0.361 \pm 0.013$ \\
SMC (4 resamples) & 3,328
& $31.4 \pm 2.0$ & $84.1 \pm 0.5$ & $88.5 \pm 1.3$
& $16.4 \pm 0.7$ & $\mathbf{0.320 \pm 0.028}$ \\
SVDD-PM          & 7,246
& $33.3 \pm 0.9$ & $\mathbf{97.4 \pm 0.1}$ & $88.2 \pm 0.0$
& $22.0 \pm 0.8$ & $0.574 \pm 0.193$ \\
TreeG-SD         & 10,000
& $31.6 \pm 0.5$ & $96.9 \pm 0.3$ & $89.1 \pm 0.3$
& $20.8 \pm 0.0$ & $0.425 \pm 0.033$ \\
DTS (Null Reward) & 15,350
& $31.1 \pm 0.8$ & $95.5 \pm 0.2$ & $89.1 \pm 0.5$
& $19.7 \pm 0.1$ & $0.337 \pm 0.036$ \\
\textbf{DTS}              & 15,450
& $\mathbf{41.6 \pm 6.8}$ & $90.4 \pm 3.2$ & $84.5 \pm 3.4$
& $\mathbf{24.1 \pm 2.2}$ & $0.331 \pm 0.158$ \\
\bottomrule
\end{tabular}%
}
\end{table}

\subsection{Symmetry-Conditioned Generation}
\label{sec:symmetry}
We test whether DTS can amplify a property-conditional model on a rare, non-differentiable target: matching a specific space group.
Both DTS and the best-of-$N$ baseline use the same space-group-conditional MatterGen variant; the reward is the indicator $r(\rvx_0) = \mathbbm{1}[\mathrm{SG}(\rvx_0) = g^\star]$, with $\mathrm{SG}(\cdot)$ computed using pymatgen \citep{ong2013python}. We focus here on space group 204 ($Im\bar{3}$), which appears in the training set with frequency $\sim$0.2\%; results for additional rare target groups are in Appendix \ref{app:symmetry}.

\cref{fig:symmetry_sg204} shows cumulative SUN $\cap$ SG 204 counts versus NFE budget. DTS produces roughly an order of magnitude more on-target SUN structures than best-of-$N$, with the gap widening as compute increases. The unconditional MatterGen, with or without DTS, generates zero structures in SG 204 across the same budget. This contrast highlights the role of each component: the conditional proposal is \emph{necessary} for DTS to access the targeted mode at all (the unconditional base places negligible mass there), but is not \emph{sufficient} to produce on-target samples at useful rates. Search compounds the proposal by reusing partial trajectories that lead into the target region.

\begin{wrapfigure}{r}{0.35\linewidth}
\vspace{-1em}
    \centering
    \includegraphics[width=\linewidth]{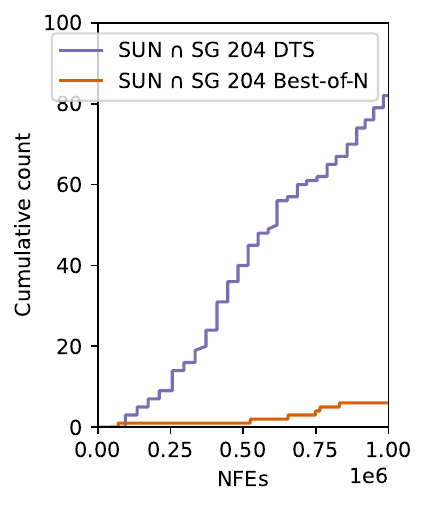}
    \caption{Cumulative count of SUN structures matching target space group 204 ($Im\bar{3}$) versus NFE budget. Both methods use the same space-group-conditional MatterGen variant.}
    \label{fig:symmetry_sg204}
    \vspace{-2em}
\end{wrapfigure}

\subsection{Multi-objective: Dielectric Materials}
\label{sec:dielectric}

We consider a multi-objective setting from \citet{riebesell2024pushing}: simultaneously maximizing bandgap and dielectric constant. These properties are approximately inversely related \citep{penn1962wave,robertson2005high,onishi2024universal}, inducing a non-convex Pareto front.
This task also demonstrates the conditional-model adaptation: MatterGen can be fine-tuned to condition jointly on (bandgap, dielectric constant), used here as a guided proposal inside DTS.

\paragraph{Setup.}
Both properties are predicted with ALIGNN \citep{choudhary2021atomistic}. We use a MatterGen base model fine-tuned on labels from this predictor, and run multi-objective DTS at 2.5M NFEs.
As a baseline, we condition MatterGen on five target points evenly spaced along the dataset Pareto front and take the union of generated samples at matched NFE budget. We evaluate after filtering to SUN structures, reporting Pareto-front size, hypervolume (HV), $R_2$ \citep{hansen1998evaluating}, and generational distance plus (GD+) \citep{ishibuchi2015modified} against the dataset Pareto front. We also include threshold counts: number of structures whose bandgap $\times$ dielectric constant exceeds fixed thresholds (30, 60, 120, 240) -- these values correspond to linear scalarization in log space.

\begin{figure}[h!]
    \centering
    \includegraphics[width=0.8\linewidth]{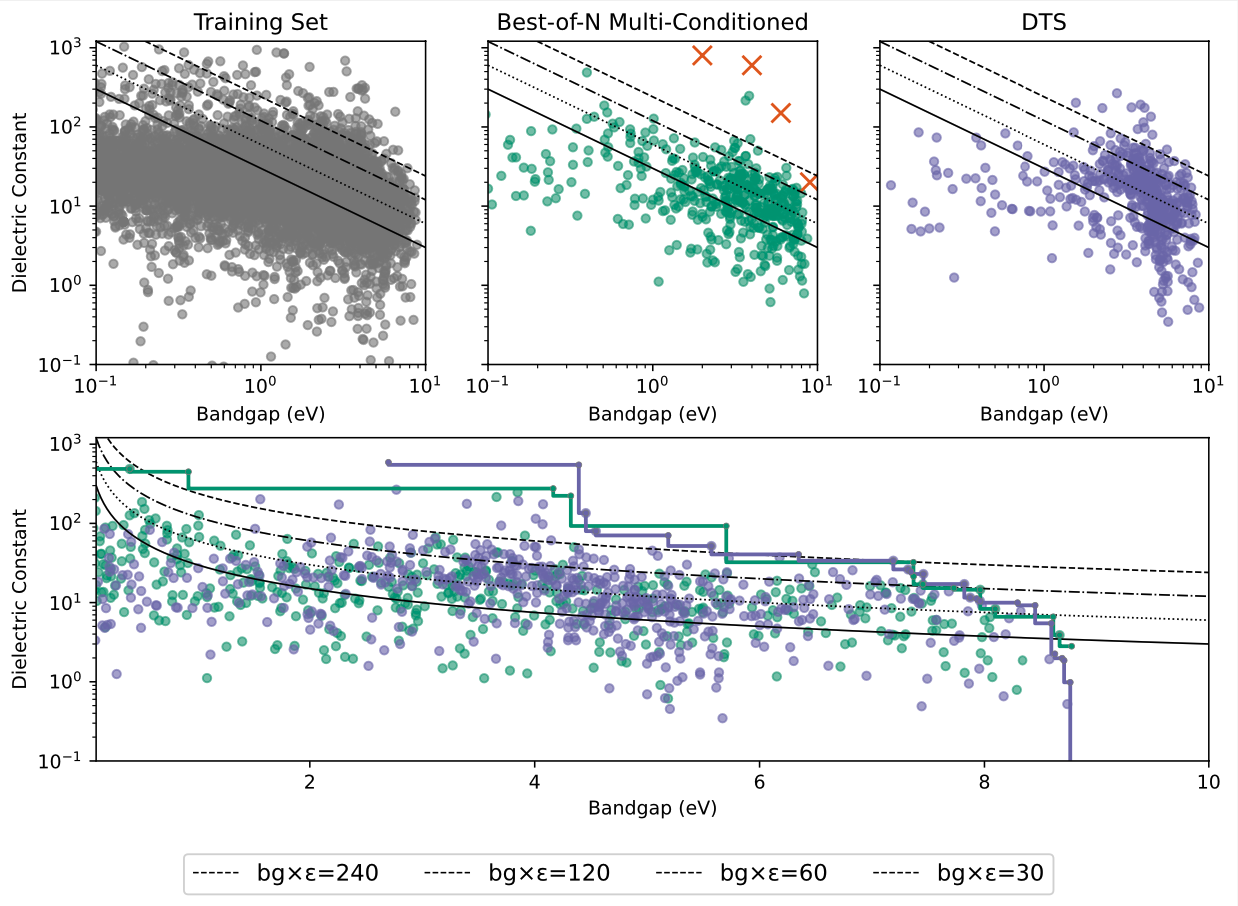}
    \caption{Dielectric constant vs.\ bandgap for SUN samples at 2.5M NFEs ($\sim$2500 total samples per method). Diagonal lines indicate constant values of bandgap $\times$ dielectric, and X markers represent conditioning targets.}
    \label{fig:dbg_scatter}
\end{figure}

\paragraph{Results.}
\cref{fig:dbg_scatter} plots SUN samples and \cref{tab:dbg_metrics} summarizes metrics.  MO-DTS  outperforms the  baseline at matched NFE budget for most metrics. 

\begin{table}[h]
\caption{Summary metrics for the bandgap vs. dielectric constant experiment. $|PF|$ = Pareto front size. Entries are mean $\pm$ std across seeds. Threshold columns count structures where bandgap\,$\times$\,dielectric $\geq$ threshold.}
\label{tab:dbg_metrics}

\resizebox{\textwidth}{!}{%
\begin{tabular}{lrrrrrrrrrr}
\toprule
 Method & $N$ & $|PF|$ & $N_{\geq 30}$ & $N_{\geq 60}$ & $N_{\geq 120}$ & $N_{\geq 240}$ & HV$\uparrow$ & GD+$\downarrow$  & R2$\downarrow$ \\
\midrule
Training set & 27,136 & 20 & 4,180 & 1,920 & 526 & 109 & 0.309 & 0.000 &  0.149 \\
MatterGen (cond.) & \textbf{665 $\pm$ 0} & 14 $\pm$ 1 & 262 $\pm$ 3 & 138 $\pm$ 6 & 28 $\pm$ 2 & 4 $\pm$ 0 & 0.118 $\pm$ 0.011 & 0.098 $\pm$ 0.024 & \textbf{0.292 $\pm$ 0.002} \\
MO-DTS & 645 $\pm$ 40 & 16 $\pm$ 1 & \textbf{402 $\pm$ 54} & \textbf{234 $\pm$ 27} & \textbf{67 $\pm$ 12} & \textbf{17 $\pm$ 4} & \textbf{0.173 $\pm$ 0.057} & \textbf{0.054 $\pm$ 0.015} & 0.338 $\pm$ 0.040 \\

\bottomrule
\end{tabular}
}
\vspace{-0.25em}
\end{table}

\subsection{Density/bulk modulus: partially aligned reward}
\label{sec:density}
The previous experiments used either unconditional models or models fine-tuned for the target reward. Conditional DTS is more general, since importance sampling decouples the proposal from the target. To show this, we repurpose an existing MatterGen model conditioned on density as a proposal, and set the target to be generating crystals with a high bulk modulus (300 GPa). In the training set the two correlate (Pearson $r = 0.512$): crystals with density $\geq 15$\,g/cm$^3$ are rare (${<}2\%$) but about $20\times$ more likely than average to have bulk modulus $\geq 300$\,GPa, so conditioning on a density of 15\,g/cm$^3$ is a useful but imperfect proxy. The mismatch is that training crystals near the bulk modulus target span a broad density range, so without correction the proposal should raise yield while distorting the density of the resulting crystals; IS should remove the distortion.

We compare four arms (\cref{tab:is-density-bulk}): DTS with an unconditional model; the density-conditioned proposal alone (best-of-$N$); conditional DTS without IS; and conditional DTS with IS. All DTS arms use reward $r(\rvx) = -\lvert K(\rvx) - 300 \rvert / 400$ with $K$ the bulk modulus predicted by ALIGNN \citep{choudhary2021atomistic} in GPa, $\lambda = 20$, $\alpha = 0.3$, and guidance strength $\gamma = 1$ for the conditional proposal; each arm runs $10^6$ NFEs over 3 seeds, with metrics computed on SUN crystals only. We report the number of SUN crystals within 30\,GPa of the target, the mean absolute error to the target, and the density distortion: the Wasserstein distance $W_1$ between the density distribution of generated crystals within 30\,GPa of the target and that of training crystals in the same range.

\begin{table}[h]
  \centering
  \caption{Results for bulk modulus experiment, ablating conditioning and importance sampling correction (IS).}
  \label{tab:is-density-bulk}
  \begin{tabular}{lccccc}
    \toprule
    Model & DTS & IS & \makecell{SUN crystals\\within 30\,GPa} $\uparrow$ & \makecell{Bulk MAE\\to 300\,GPa} $\downarrow$ & \makecell{Near-target density\\$W_1$ to training} $\downarrow$ \\
    \midrule
    Unconditional       & \greencheck &      \redx        & 237 & 108.23 & 2.80 \\
    Density-Conditioned & \redx & \redx  & 61  & 131.02 & 4.72 \\
    Density-Conditioned & \greencheck & \redx        & \textbf{302} & \textbf{62.82}  & 4.31 \\
    Density-Conditioned & \greencheck & \greencheck       & 260 & 70.33  & \textbf{2.15} \\
    \bottomrule
  \end{tabular}
\vspace{-0.25em}
\end{table}

The mismatched proposal is useful only together with search: on its own it yields four times fewer near-target SUN crystals than unconditional DTS, since it shifts samples toward dense crystals but does not select for bulk modulus. Conditional DTS without IS outperforms both, but the proposal creates biased samples: the resulting near-target crystals are much denser than training crystals in the same range. Importance correction removes this bias while keeping most of the gain in yield. An existing conditional model can therefore serve as a proposal for a new reward, with the output governed by the reward rather than by the proposal.

\section{Related Work}
\label{sec:related}

\paragraph{Inference-time steering of diffusion models.}
Score-modifying approaches include classifier-based \citep{dhariwal2021diffusion} and classifier-free \citep{ho2022classifier} guidance; both approximate reward-tilted sampling but are biased and require differentiable rewards. Particle-based methods treat denoising as SMC \citep{singhal2025general}. 
Methods such as SVDD \citep{li2024derivative} and TreeG \citep{guo2026training} perform a local search: they  branch each denoising step into several candidates and keep the best by a value estimate.
The closest line to us organizes intermediate states into a tree: Diffusion Tree Sampling \citep{jain2025diffusion} formalizes this for continuous diffusion as MCTS with soft-Bellman backups and is the basis for our method. PepTune \citep{tang2025peptune} develops Monte Carlo Tree Guidance for discrete diffusion in multi-objective peptide design with a Pareto-dominance reward. Our approach operates on continuous diffusion with scalarization-based per-objective backups, and composes with conditional  models via importance correction. A separate strand fine-tunes for specific objectives via conditional training \citep{zeni2023mattergen} or RL \citep{black2023training,chen2025matinvent}; our work leaves the model fixed and composes with such fine-tuned variants when available.

\paragraph{Multi-objective generation.}
Multi-objective generation has a long history in evolutionary algorithms and Bayesian optimization \citep{daulton2020differentiable}; multi-objective GFlowNets \citep{jain2023multi, zhu2023sample} sample from a Pareto-tilted distribution by conditioning on preferences during training. Diffusion-based methods modify the score or velocity field during a single denoising pass (e.g. PROUD \citep{yao2024proud}, ParetoFlow \citep{yuan2024paretoflow}, and Preference-Guided Diffusion \citep{annadani2025preference}), use diffusion as proposals \citep{kong2024diffusion}, or wrap diffusion in outer BO loops \citep{gruver2023protein}. Our method differs by using tree-based search over pretrained continuous diffusion models, using per-objective values and backups, yielding the target of \cref{eq:pareto_mixture} and admitting importance-corrected use of conditional models.

\section{Discussion}
We adapted Diffusion Tree Sampling to crystal generative models in two ways: importance-corrected guided proposals, which let existing property-conditional models steer the search without changing its target, and a multi-objective extension, in which per-objective value backups and a preference sampled per traversal let a single tree search over the Pareto front of competing properties.
DTS alone improved stability and scaled with compute, beating best-of-$N$, but could not reach rare regions in the space of possible materials.
This access can be provided by using conditional proposals, but at the cost of tying the output to whatever the proposal was tuned for.
This dependence can be solved via importance correction, enabling cases where the proposal and reward are mismatched, or where we want a single tree to capture multiple different preferences.

The main caveat is reward fidelity. All reported rewards use learned surrogates, so search can amplify systematic predictor errors; DFT validation and synthesis remain essential next steps. Importance correction is also most reliable when guided and base models share a backbone; separately trained proposals may require stronger weight control.

More broadly, this method fits the foundation-model setting: a pretrained generative prior, adapted at inference time to objectives that it has not been specifically trained for.
This setup benefits from inference-time search, where more compute is spent per candidate to produce higher-quality samples. Materials discovery is a natural fit because generation is cheap next to DFT validation and synthesis, but the same technique could also be applied to other AI for science settings, such as  protein and small-molecule design.

\section*{Acknowledgment}
This project is in part funded by CIFAR AI chairs, NSERC Discovery grant and IVADO.
D.L.'s research is additionally supported by FRQNT.
Computational resources are provided by Mila and the Digital Research Alliance of Canada.

\bibliography{bib}
\bibliographystyle{plainnat}

\newpage
\appendix
\onecolumn

\section{Algorithm}
\label{app:algo}
\begin{algorithm}[H]
\caption{Diffusion Tree Sampling with guided proposals and multiple objectives.
Unmarked lines are DTS \citep{jain2025diffusion}; lines tagged \textcolor{isblue}{[IS]} add the
importance-corrected guided proposals of \cref{sec:cond}, and lines tagged \textcolor{mored}{[MO]} add
multi-objective selection and backup of \cref{sec:modts}. Batched execution is described in \cref{sec:parallel}.}
\label{alg:dts-annotated}
\begin{algorithmic}[1]
\STATE \textbf{Input:} base model $p_\theta$; inverse temperature $\lambda$;  NFE budget $N_{\mathrm{NFE}}$ ;
\STATE \textbf{Input:}
widening parameters $C$,$\alpha$, branching timesteps $\mathcal{T}$
\STATE \textbf{Input:}
\is{guided proposal $q_c$} (default $q_c=p_\theta$) \tIS
\STATE \textbf{Input:}
\mo{objectives $\rvr=(r_1,\dots,r_m)$; scalarization $s$; preferences $p(w)$; condition map $w\mapsto c_w$} (default $m=1$, $s=\mathrm{id}$, $c_w\equiv c$) \tMO
\STATE Initialize dummy root $\rvx_{T+1}$ with $p(\rvx_T\mid\rvx_{T+1})=p_T$;\quad $\hat{\rvv}(\rvx_{T+1})\gets\mathbf0$, $N(\rvx_{T+1})\gets1$, $\mathcal{C}(\rvx_{T+1})\gets\emptyset$
\STATE $\mathrm{nfe}\gets0$
\WHILE{$\mathrm{nfe}<N_{\mathrm{NFE}}$}
 \cmt{Initialize path}
  \STATE $P\gets\{\rvx_{T+1}\}$;\quad $t\gets T+1$
  \STATE \mo{$w\sim p(w)$;\quad $c\gets c_w$} \tMO
  \cmt{Selection: descend until reaching a terminal or an expandable node}
\WHILE{$t>0$}
  \IF{$t\in\mathcal T$ \textbf{and} $|\mathcal C(\rvx_t)|<C\,N(\rvx_t)^{\alpha}$} \STATE \textbf{break} \ENDIF
    \STATE $\rvx_{t-1}\gets\rvz\in\mathcal{C}(\rvx_t)$ with probability
      $\dfrac{\is{\eta(\rvz)}\,\exp\!\big(\lambda\,\mo{s(\hat{\rvv}(\rvz) \mid w)}\big)}
            {\sum_{\rvz'\in\mathcal{C}(\rvx_t)}\is{\eta(\rvz')}\,\exp\!\big(\lambda\,\mo{s(\hat{\rvv}(\rvz') \mid w)}\big)}$ \tISMO
    \STATE $P\gets P\cup\{\rvx_{t-1}\}$;\quad $t\gets t-1$
  \ENDWHILE
  \cmt{Expansion and rollout: grow a new path to $t=0$, adding every node to the tree}
  \WHILE{$t>0$}
    \STATE $\rvx_{t-1}\sim\is{q_c}(\,\cdot\mid\rvx_t)$;\quad $\mathrm{nfe}\gets\mathrm{nfe}+1$ \tIS
    \STATE $\hat{\rvv}(\rvx_{t-1})\gets\mathbf0\in\mo{\sR^m}$;\quad $N(\rvx_{t-1})\gets1$;\quad $\mathcal{C}(\rvx_t)\gets\mathcal{C}(\rvx_t)\cup\{\rvx_{t-1}\}$ \tMO
    \STATE \is{$\eta(\rvx_{t-1})\gets p_\theta(\rvx_{t-1}\mid\rvx_t)\,/\,q_c(\rvx_{t-1}\mid\rvx_t)$} \tIS
    \STATE $P\gets P\cup\{\rvx_{t-1}\}$;\quad $t\gets t-1$
  \ENDWHILE
  \cmt{Terminal evaluation}
  \STATE $\hat{\rvv}(\rvx_0)\gets\mo{\rvr(\rvx_0)}$ \tMO
  \cmt{Backup: per-objective self-normalized soft Bellman update}
  \FOR{each $\rvx_t\in P$ with $t\ge1$, from $\rvx_1$ up to $\rvx_{T+1}$}
    \STATE \mo{\textbf{for} $i=1,\dots,m$:}\quad
      $\hat v_i(\rvx_t)\gets\dfrac1\lambda\log
       \dfrac{\sum_{\rvz\in\mathcal{C}(\rvx_t)}\is{\eta(\rvz)}\,e^{\lambda\hat v_i(\rvz)}}
             {\sum_{\rvz\in\mathcal{C}(\rvx_t)}\is{\eta(\rvz)}}$ \tISMO
    \STATE $N(\rvx_t)\gets N(\rvx_t)+1$
  \ENDFOR
\ENDWHILE
\RETURN terminal nodes $\{\rvx_0\}$ of the tree
\end{algorithmic}
\end{algorithm}

\section{Connection to Basin Hopping and Monte Carlo Minimization}
\label{app:basin}
DTS bears a structural resemblance to a family of classical global optimization methods used in computational chemistry, most notably Monte Carlo Minimization \citep{li1987monte} and its successor, basin hopping \citep{wales1997global,wales1999global}. These algorithms iterate a perturb–relax–accept cycle: a candidate configuration is randomly displaced, mapped to the bottom of the resulting basin via a deterministic local minimizer, and accepted or rejected by a Metropolis criterion at temperature $T$ on the minimized energies. 

Different components of DTS have natural counterparts in this framework. Tree expansion of a node $x_t$ plays the role of perturbation, with the noise level $t$ controlling an effective step size; the unrolled reverse process from $t$ to $0$ plays the role of local minimization, an interpretation supported by the established equivalence between denoising score matching and learning a Boltzmann force field over equilibrium structures \citep{zaidi2022pre,arts2023two}; the soft Bellman backup with inverse temperature $\lambda$ plays the role of Metropolis acceptance, with the greedy variant of DTS recovering monotonic basin hopping.

Several features distinguish our method from these classical methods.
The search tree retains and shares value estimates across all visited nodes rather than maintaining a single chain.
Rather than just minimizing energy via structural relaxation, using a generative model enables swapping of atom types in a structure. 
Lastly, the equivalence between score matching and force fields does not apply to property-conditional models, which steer generation towards specific properties rather than just energy minima.

\section{Illustrative experiments}
\label{app:toy}

\subsection{Gaussian-mixture toy illustration of the double-tilt problem}

\begin{figure}[h]
    \centering
    \includegraphics[width=\linewidth]{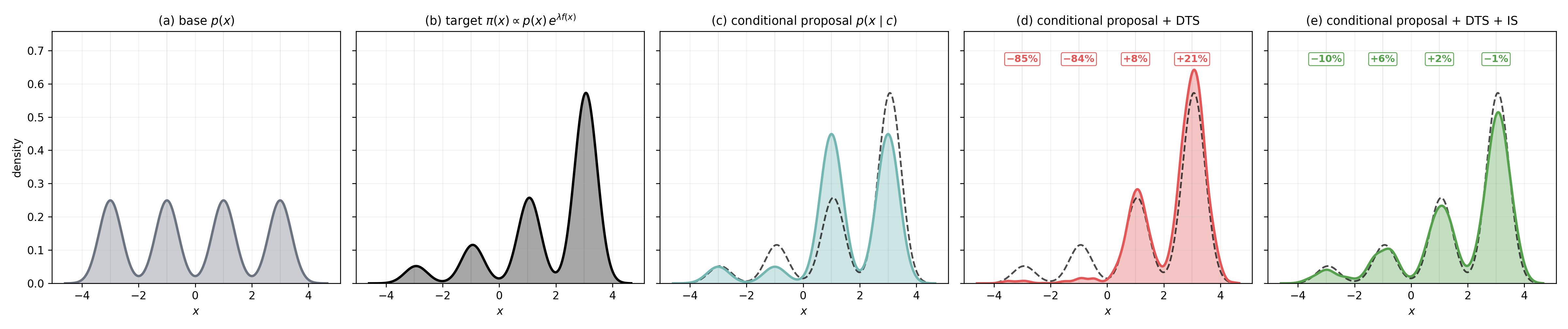}
    \caption{(a) The unconditional distribution $p(\mathbf{x})$ is an equal-weight four-component Gaussian mixture; (b) with linear reward $f(x) = x$, the analytic target $\pi(x) \propto p(\mathbf{x})\,\exp(\lambda f(x))$ concentrates mass on higher-$\mathbf{x}$ modes; (c) the conditional proposal $p(\mathbf{x} | c)$ is biased toward positive $\mathbf{x}$ and aligns better with the target (dashed black); (d) naively running DTS with $p(\mathbf{x} | c)$ as the proposal double-tilts the distribution: the two left modes collapse by roughly $85\%$ and the rightmost mode overshoots by $21\%$; (e) multiplying the reward weights by the importance ratio $p(\mathbf{x})/p(\mathbf{x}|c)$ recovers $\pi$ to within a few percent at every mode.}
    \label{fig:gaussian_toy}
\end{figure}

\cref{fig:gaussian_toy} illustrates the double-tilt problem and the importance correction of \cref{sec:cond} on a Gaussian mixture. The unconditional distribution is an equal-weight four-component mixture; with linear reward $r(x)=x$ the analytic target $\pi(x)\propto p(\rvx)\exp(\lambda f(x))$ concentrates on the higher-$x$ modes. A conditional proposal $p(\rvx\mid c)$ biased toward positive $x$ aligns better with $\pi$, but naively running DTS with this proposal double-tilts and distorts the distribution. Multiplying the reward weights by the importance ratio recovers $\pi$ to within a few percent at every mode.

\subsection{Illustration on a Gaussian mixture in multi-objective setting}

\begin{figure}[h]
    \centering
    \includegraphics[width=\linewidth]{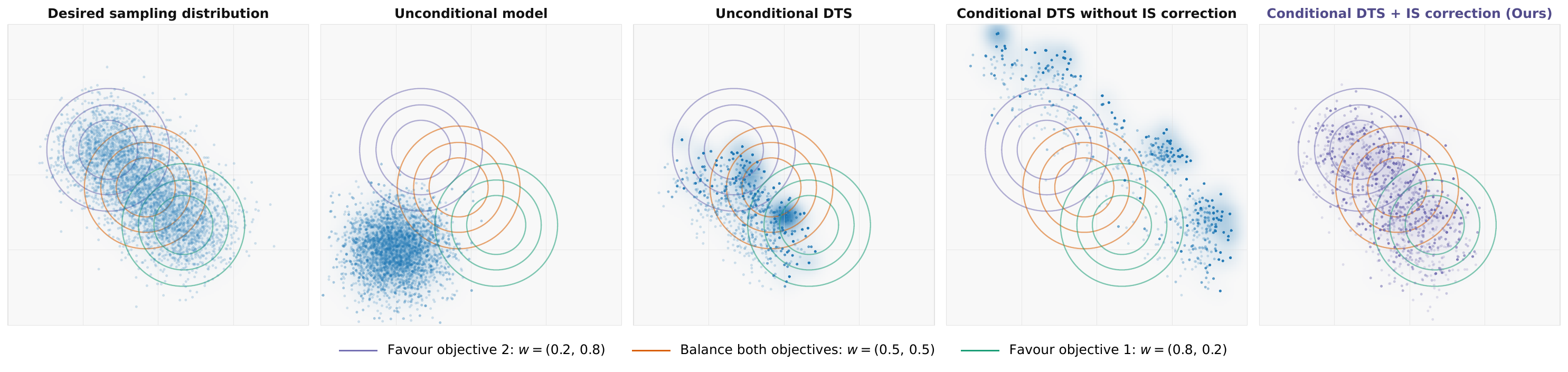}
    \caption{(a) The target distribution is a mixture of different preferences over the tilted distribution $\pi(x,y) \propto p(x,y)\,\exp(w_1 x + w_2 y)$; (b) the unconditional distribution $p(x,y)$ is a two-dimensional standard Gaussian; (c) using DTS with the unconditional proposal fails to sample accurately from the target distribution due to low density of the base distribution in this region; (d) naively running DTS with a conditional model $p(x,y|c)$ as the proposal double-tilts the distribution; (e) multiplying the reward weights by the importance ratio $p(x,y)/p(x,y|c)$ produces samples that align better with the target distribution.}
    \label{fig:multi_gaussian_toy}
\end{figure}

\Cref{fig:multi_gaussian_toy} illustrated how our method handles multiple conflicting objectives by scalarizing different preferences over the objectives in a single shared tree. The unconditional distribution is a two-dimensional Gaussian $p(x,y) = \mathcal{N}((0,0); I)$. The two objectives $f_1(x,y) = x$ and $f_2(x,y) = y$ with corresponding weights $w_1$ and $w_2$ tilt the distribution to give the target $\pi(x,y) \propto p(x,y)\,\exp(w_1 x + w_2 y)$. Different preferences $\mathbf{w} = (w_1, w_2)$ define different target distributions, and our aim is to produce samples from the Pareto front that covers all possible preferences. For illustrative purposes, we fix three specific preferences.

The unconditional model, even with DTS, fails to produce samples from the target distribution since the proposal model does not have high mass in the target region. We use a conditional model that is conditioned on $(w_1 x + w_2 y)$ to produce samples closer to the target region and use it as a guided proposal within DTS. However, naively replacing the proposal results in \emph{double-tilting}, so the samples lie outside the target distribution. We fix this by using an importance sampling correction (as described in \Cref{sec:cond}) that accounts for the conditional model. 

\section{Predictor--corrector and other multi-stage samplers}
\label{app:pc}

The discussion in \cref{sec:cond} assumes a single-step reverse transition. In practice, modern diffusion samplers often implement a \emph{multi-stage} update between $t$ and $t-1$, for improved stability and sample quality. A common pattern is \emph{predictor--corrector} (PC) sampling, where a predictor step performs a discretized reverse-SDE/ODE update, and a corrector step applies one or more refinement moves (e.g., Langevin/MCMC steps) that approximately preserve the desired marginal at the new noise level. More broadly, modern samplers may use higher-order solvers, adaptive step sizes, or alternating deterministic/stochastic updates; our method remains valid as long as we can write down (or evaluate up to a constant) the probability of the composite transition used by the base sampler.

To incorporate PC sampling into guided proposals and importance correction, we treat the overall $t\to t-1$ move as a composition through an intermediate state $\tilde{\rvx}_{t-1}$:
\begin{equation*}
p_\theta(\rvx_{t-1}\mid \rvx_t) = \int p_{\text{pred}}(\tilde{\rvx}_{t-1}\mid \rvx_t)\,p_{\text{corr}}(\rvx_{t-1}\mid \tilde{\rvx}_{t-1})\,d\tilde{\rvx}_{t-1},
\end{equation*}
with an analogous guided/conditional composite transition where $q_{\text{pred},w}(\cdot\mid \rvx_t):=p_{\text{pred}}(\cdot\mid \rvx_t, c(w))$ and similarly for $q_{\text{corr},w}$. In the tree expansion/rollout we explicitly sample $\tilde{\rvx}_{t-1}$ and then $\rvx_{t-1}$, and use a factorized importance ratio consistent with this sampling path:
\begin{equation}
\frac{p_\theta(\rvx_{t-1}\mid \rvx_t)}{q_w(\rvx_{t-1}\mid \rvx_t)}
\;\approx\;
\frac{p_{\text{pred}}(\tilde{\rvx}_{t-1}\mid \rvx_t)}{q_{\text{pred},w}(\tilde{\rvx}_{t-1}\mid \rvx_t)}
\;\times\;
\frac{p_{\text{corr}}(\rvx_{t-1}\mid \tilde{\rvx}_{t-1})}{q_{\text{corr},w}(\rvx_{t-1}\mid \tilde{\rvx}_{t-1})}.
\label{eq:pc_ratio}
\end{equation}
This replaces the single-step ratio in \cref{eq:is_backup}. Although the right-hand side is not the marginal density ratio for $\rvx_{t-1}$ alone, it is the exact path-space importance weight for the joint augmented transition $(\tilde\rvx_{t-1},\rvx_{t-1})$ actually sampled by the algorithm; the $\approx$ symbol reflects that the marginalization over $\tilde\rvx_{t-1}$ is performed implicitly by sampling. In particular, if the corrector is shared between conditional and unconditional samplers (i.e., $q_{\text{corr},w}=p_{\text{corr}}$), the corrector ratio cancels and only the predictor ratio is needed. When both predictor and corrector differ (e.g., guided score used in both stages), \cref{eq:pc_ratio} yields the appropriate weight for maintaining consistency with the unconditional-base target while still leveraging guided proposals during the full sampling procedure.


\section{Scalarizations}
\label{app:scalarization}

As explained in \Cref{sec:modts}, a standard way to operationalize multi-objective optimization is \emph{scalarization}: decompose it into a family of single-objective problems of the form $\max_{\rvx}\, r(\rvx\mid w)$, where $w \in \Delta^{m-1}\coloneqq\{w\in\mathbb{R}^m_{\ge 0}:\sum_i w_i=1\}$ encodes preferences and $r(\rvx \mid w) \coloneqq s(\rvr(\rvx) \mid w)$ for scalarization $s$. The simplest choice for $s$ is the weighted sum, $s(\rva \mid w) = \sum_{i=1}^m w_i\, a_i$. An important alternative is (weighted) \emph{Tchebycheff} scalarization \citep{bowman1976relationship}, which measures the worst (weighted) deviation from an \emph{ideal} vector $\rvz \in\mathbb{R}^m$:
\begin{equation}
s(\rva \mid w,\rvz) \;=\; - \max_{i \in \{1,\dots,m\}} w_i\,\abs{a_i-z_i},
\end{equation}
i.e., maximizing $s(\cdot\mid w,\vz)$ is equivalent to minimizing the worst weighted distance to the ideal. This encourages balanced progress across objectives and can recover Pareto-optimal points that may be missed by linear scalarization on non-convex fronts. Both linear and Tchebycheff scalarizations can be unified via a generalized $p$-norm form,
\begin{equation}
s\!\left(\rva\mid w,\vz\right) \;=\; -\norm{ w \odot \abs{\rva -\rvz} }_{p},
\end{equation}
where $\odot$ is elementwise multiplication. Linear and Tchebycheff scalarizations are recovered for $p=1$ and $p=\infty$ respectively. More generally, other weight-parameterized scalarizations (e.g., augmented Tchebycheff, PBI, achievement scalarizing functions) can be viewed as different ways to sweep the Pareto front. Our method is \emph{indifferent} to the particular scalarization: any mapping $w \mapsto s(\cdot\mid w)$ is compatible as long as we can systematically choose $w$ from a specified set/distribution and evaluate the resulting scalar objective.

\section{Properties of the per-objective value surrogate}
\label{app:approx}

Scalarizing only at selection time enables computation sharing across all preferences. The exact soft value of the scalarized reward, $v_t^\star(\rvx_t\mid w) := \tfrac{1}{\lambda}\log\E[\exp(\lambda\, r(\rvx_0\mid w))\mid \rvx_t]$, would instead require a separate backup tree for every $w$. The per-objective surrogate of $v_t(\rvx_t | w)$ relates to it as follows.

\begin{proposition}
\label{prop:approx}
Let $v_t^\star(\rvx_t\mid w)$ be the exact scalarized soft value, and let
\begin{equation}
\label{eq:approx_soft_value}
v_{t,i}(\rvx_t) \;:=\; \tfrac{1}{\lambda}\log\E_{p_\theta(\rvx_0\mid \rvx_t)}\!\big[\exp(\lambda\, r_i(\rvx_0))\big],
\end{equation}
be the per-objective values used by MO-DTS, with $v_t(\rvx_t | w) \coloneqq s(\rvv_t(\rvx_t) | w)$ its scalarization. Assume the rewards $r_i$ are bounded. Then:
\begin{enumerate}
\item For \emph{linear} scalarization, $v_t^\star(\rvx_t\mid w)\le v_t(\rvx_t\mid w)$ for every $\lambda>0$.
\item For any continuous scalarization that is non-decreasing in each objective, $v_t^\star$ is upper bounded by $v_t$ in the low-temperature limit:
\begin{equation}
\lim_{\lambda\to\infty}v_t^\star(\rvx_t\mid w)
\;\le\;
\lim_{\lambda\to\infty}v_t(\rvx_t\mid w).
\end{equation}
\item For any continuous scalarization, the approximation becomes exact at $t=0$:
\begin{equation}
v_0(\rvx_0\mid w)=v_0^\star(\rvx_0\mid w),
\end{equation}
and, whenever the conditional uncertainty in $\rvr(\rvx_0)$ given $\rvx_t$ vanishes as $t\to0$, for every fixed $\lambda>0$,
\begin{equation}
\abs{v_t(\rvx_t\mid w)-v_t^\star(\rvx_t\mid w)}\to 0.
\end{equation}
\end{enumerate}
\end{proposition}

\begin{proof}
Fix $\rvx_t$ and let all expectations and essential suprema be under
$p_\theta(\rvx_0\mid \rvx_t)$. Write
$R_i:=r_i(\rvx_0)$ and $\rvr:=(R_1,\dots,R_m)$.

\paragraph{(i) Linear scalarization, all $\lambda$.}
Assume $s(\rva\mid w)=\sum_i w_i a_i$ with $w_i\ge 0$ and $\sum_i w_i=1$, and set
$X_i:=\exp(\lambda R_i)$. Then
\begin{equation}
\E[\exp(\lambda s(\rvr\mid w))]
\;=\; \E\!\Big[\textstyle\prod_i X_i^{w_i}\Big].
\end{equation}
H\"older's inequality, with conjugate exponents $1/w_i$ for the positive weights, gives
\begin{equation}
\E\!\Big[\textstyle\prod_i X_i^{w_i}\Big]
\;\le\; \textstyle\prod_i \E[X_i]^{w_i}.
\label{eq:proof_holder}
\end{equation}
Taking $\tfrac{1}{\lambda}\log$ on both sides yields
\begin{equation}
v_t^\star(\rvx_t\mid w)
\;\le\; \sum_i w_i\,\tfrac{1}{\lambda}\log\E[X_i]
\;=\; \sum_i w_i\, v_{t,i}(\rvx_t)
\;=\; v_t(\rvx_t\mid w).
\end{equation}
Equality in \cref{eq:proof_holder} holds iff the active tilted variables are almost surely proportional, equivalently $R_i-R_j$ is almost surely constant for all pairs with $w_i,w_j>0$.

A second-order Taylor expansion of the cumulant generating function gives, for bounded rewards,
\begin{align}
v_{t,i}(\rvx_t)
&= \E[r_i\mid \rvx_t]
 + \tfrac{\lambda}{2}\,\mathrm{Var}(r_i\mid \rvx_t)
 + O(\lambda^2),\\
v_t^\star(\rvx_t\mid w)
&= \E[r(\rvx_0\mid w)\mid \rvx_t]
 + \tfrac{\lambda}{2}\,\mathrm{Var}(r(\rvx_0\mid w)\mid \rvx_t)
 + O(\lambda^2).
\end{align}
Under linear scalarization, $\E[r(\rvx_0\mid w)\mid \rvx_t]=\sum_i w_i \E[r_i\mid \rvx_t]$, so the zeroth-order terms agree and
\begin{equation}
v_t(\rvx_t\mid w) - v_t^\star(\rvx_t\mid w)
\;=\; \tfrac{\lambda}{2}\Big(\sum_i w_i\,\mathrm{Var}(r_i\mid \rvx_t) - \mathrm{Var}(\textstyle\sum_i w_i r_i\mid \rvx_t)\Big) + O(\lambda^2),
\end{equation}
whose first-order coefficient is non-negative by convexity of variance. The exact non-negativity for all $\lambda$ is the H{\"o}lder bound above.

\paragraph{(ii) Monotone scalarizations, low temperature.}
We use the standard log-sum-exp limit: if a random variable $Y$ is bounded above, then
\begin{equation}
\lim_{\lambda\to\infty}\frac{1}{\lambda}\log\E[\exp(\lambda Y)]
\;=\;
\operatorname*{ess\,sup} Y.
\label{eq:proof_laplace}
\end{equation}
Indeed, the upper bound follows from $\E e^{\lambda Y}\le e^{\lambda\operatorname*{ess\,sup}Y}$; for any $\epsilon>0$, the event $Y\ge \operatorname*{ess\,sup}Y-\epsilon$ has positive probability, giving the matching lower bound after dividing by $\lambda$ and taking $\lambda\to\infty$.

Applying \cref{eq:proof_laplace} to $Y=s(\rvr\mid w)$ gives
\begin{equation}
\lim_{\lambda\to\infty} v_t^\star(\rvx_t\mid w)
\;=\;
\operatorname*{ess\,sup} s(\rvr\mid w).
\end{equation}
Applying it componentwise gives $v_{t,i}(\rvx_t)\to \operatorname*{ess\,sup}R_i$. By continuity of $s(\cdot\mid w)$,
\begin{equation}
\lim_{\lambda\to\infty}v_t(\rvx_t\mid w)
\;=\;
s\!\big((\operatorname*{ess\,sup}R_1,\dots,\operatorname*{ess\,sup}R_m)\mid w\big).
\end{equation}
Since $R_i\le \operatorname*{ess\,sup}R_i$ almost surely for every $i$, coordinatewise monotonicity implies
$s(\rvr\mid w)\le s\!\big((\operatorname*{ess\,sup}R_1,\dots,\operatorname*{ess\,sup}R_m)\mid w\big)$ almost surely. Therefore
\begin{equation}
\operatorname*{ess\,sup} s(\rvr\mid w)
\;\le\;
s\!\big((\operatorname*{ess\,sup}R_1,\dots,\operatorname*{ess\,sup}R_m)\mid w\big),
\end{equation}
which proves the claimed low-temperature optimistic bound.

\paragraph{(iii) Low-noise limit.}
At $t=0$, the conditional law $p_\theta(\rvx_0'\mid \rvx_0)$ is a point mass at $\rvx_0$, so
\begin{equation}
v_{0,i}(\rvx_0)=r_i(\rvx_0),
\qquad
v_0(\rvx_0\mid w)=v_0^\star(\rvx_0\mid w)=s(\rvr(\rvx_0)\mid w).
\end{equation}
For the limiting statement, consider any sequence of low-noise conditional laws for which
$\rvr\to\bar\rvr$ in probability and the rewards remain bounded. By dominated convergence,
\begin{equation}
\E[\exp(\lambda R_i)] \to \exp(\lambda \bar r_i),
\qquad
\E[\exp(\lambda s(\rvr\mid w))] \to \exp(\lambda s(\bar\rvr\mid w)),
\end{equation}
for every fixed $\lambda>0$. Thus $v_{t,i}(\rvx_t)\to \bar r_i$ and
$v_t^\star(\rvx_t\mid w)\to s(\bar\rvr\mid w)$. Continuity of $s(\cdot\mid w)$ then gives
$v_t(\rvx_t\mid w)=s(\rvv_t(\rvx_t)\mid w)\to s(\bar\rvr\mid w)$, so
$\abs{v_t(\rvx_t\mid w)-v_t^\star(\rvx_t\mid w)}\to 0$.
\end{proof}

\paragraph{Remark on nonlinear scalarizations.}
The low-temperature statement above is intentionally asymptotic. For general nonlinear scalarizations,
$v_t^\star(\rvx_t\mid w)\le v_t(\rvx_t\mid w)$ need not hold at every finite $\lambda$. The monotonicity
condition is also essential: negative-distance Tchebycheff scalarization,
$s(\rva\mid w,z)=-\max_i w_i\abs{a_i-z_i}$, is monotone only on domains where each objective lies on the same side
of the ideal point $z_i$ and improvement means moving toward $z_i$. If feasible rewards can overshoot the
ideal, the componentwise essential supremum may move away from the ideal, and the optimistic bound can fail.
In applications that use such scalarizations, the proposition applies either after restricting to the monotone
region or after redefining the backed-up coordinates as monotone improvement scores relative to the target.

\section{Additional Experimental Details and Results}
\label{app:experiments}

\subsection{DiffCSP}
\label{app:experiments_diffcsp}

 For the DiffCSP experiment, we sweep progressive-widening parameters $\alpha$, C, and inverse temperature $\lambda$ at a 1M-NFE budget, selecting the configuration that maximizes the stability metric. The chosen configuration — $\alpha=0.3$, $C=0.5$ and $\lambda=10$ — is then evaluated at all reported NFE budgets. The batch size is set to 50.  Novelty is computed against DiffCSP's MP-20 training set. 

 We compare against SMC \citep{singhal2025general}, SVDD-PM \citep{li2024derivative}, and TreeG-SD \citep{guo2026training}, which are each training-free and gradient-free steering methods for diffusion models.
 SVDD-PM branches each particle into $10$ next-state candidates and scores their Tweedie $\hat{x}_0$ estimates. TreeG-SD samples $10$  perturbed clean targets and moves toward the highest-reward candidate. In our DiffCSP adaptation, TreeG-SD perturbs fractional coordinates and lattice variables while holding atom types fixed, since Gaussian perturbations of discrete type logits would produce arbitrary compositions. Because Tweedie estimates at high noise levels generally do not represent physically meaningful crystals, we activate branching after certain noise levels: 
$t \leq 900$ for SVDD-PM and  $t \leq 600$ for TreeG-SD, where $t$  decreases from 1000 (most noise) to 0 (no noise). SMC instead uses full terminal rollouts to calculate its resampling potentials.
 SMC is run with 400 particles over 1000 denoising steps, resampling every 20 steps for $t<990$. At each resample point, the rewards are estimated by a full rollout to $t=0$, resulting in $\approx$ 10M NFEs for rollout-based reward estimation. All reported NFE budgets fall within 2 \% of Best-of-N's nominal 10M-NFE budget.

\subsection{Wall-clock time comparisons}
\label{app:timing}
In \cref{tab:wall_clock_diffcsp}, we compare the wall-clock times for generating samples for all methods at $\approx$ 10M NFEs. For DiffCSP, the overhead is mostly from reward computation since the denoising itself is very fast. In \cref{tab:wall_clock_mattergen}, we compare the wall-clock times for 1M NFEs using MatterGen. MatterGen is much slower, so the reward calculation time makes less of a difference.

 \begin{table}[t]
  \centering
  \caption{Approximate wall-clock times on a single A100 for the 10M NFE budget for DiffCSP, with a batch size of 50.}
  \label{tab:wall_clock_diffcsp}
  \begin{tabular}{@{}lcc@{}}
  \toprule
  Method & Reward & Time \\
  \midrule
  DTS                     & -$\Ehull$        & $\approx$\,5.9\,h \\
  DTS                     & 0                & $\approx$\,5.2\,h \\
  Best-of-N baseline       & 0                       & $\approx$\,1.5\,h \\
  SMC                 & -$\Ehull$              & $\approx$\,2.2\,h \\
  \bottomrule
  \end{tabular}
  \end{table}

 \begin{table}[t]
  \centering
  \caption{Approximate wall-clock times on a single RTX8000 for the 1M NFE budget for MatterGen, with a batch size of 50.}
  \label{tab:wall_clock_mattergen}
  \begin{tabular}{@{}lcc@{}}
  \toprule
  Method & Reward & Time \\
  \midrule
  DTS                     & -$\Ehull$        & $\approx$\,3.2\,h \\
  DTS                     & 0                & $\approx$\,3.0\,h \\
  Best-of-N baseline       & 0                       & $\approx$\,2.7\,h \\
  \bottomrule
  \end{tabular}
  \end{table}

\subsection{Matched-output count results}
Due to trajectory reuse, DTS produces more candidate structures than regular best-of-N sampling. As such, we present results for the stability experiment of \cref{sec:stability} at a fixed number of output samples, rather than presenting results over all samples created using the same NFEs.

\paragraph{MatterGen.}
All methods are evaluated at  $N=2500$ terminal candidates, matching the baseline. Two selection rules are shown: first-$N$
 by NFE (stopping the search early) and best-$N$
 by predicted $\Ehull$.
\begin{table}[t]
\centering
\caption{MatterGen results with matched output counts ($N=2500$), using first-$N$ and best-$N$ selection.}
\label{tab:mattergen_matched_outputs}
\resizebox{\textwidth}{!}{%
\begin{tabular}{cllrrrr}
\toprule
$\alpha$ & Method & Selection & \# SUN & Stable (\%)
& SUN unique systems & Avg.\ $\Ehull$/atom \\
\midrule
--- & MatterGen baseline & All 2500
& $868.0$ & $68.4$ & $\mathbf{848.3}$ & $0.0858$ \\
\midrule
0.1 & DTS (Null-Reward) & First-$N$
& $733.0 \pm 26.8$ & $66.9 \pm 1.3$
& $544.0 \pm 25.2$ & $0.0904 \pm 0.0038$ \\
0.3 & DTS (Null-Reward) & First-$N$
& $789.3 \pm 32.7$ & $65.7 \pm 1.5$
& $673.7 \pm 17.6$ & $0.0911 \pm 0.0039$ \\
0.5 & DTS (Null-Reward) & First-$N$
& $813.3 \pm 24.0$ & $66.7 \pm 1.4$
& $749.7 \pm 23.4$ & $0.0901 \pm 0.0035$ \\
\addlinespace
0.1 & DTS + $\Ehull$ reward & First-$N$
& $734.3 \pm 52.2$ & $\mathbf{72.7 \pm 0.7}$
& $533.0 \pm 19.1$ & $\mathbf{0.0792 \pm 0.0007}$ \\
0.3 & DTS + $\Ehull$ reward & First-$N$
& $866.0 \pm 35.5$ & $70.6 \pm 0.9$
& $714.7 \pm 27.8$ & $0.0843 \pm 0.0021$ \\
0.5 & DTS + $\Ehull$ reward & First-$N$
& $\mathbf{921.0 \pm 22.5}$ & $69.3 \pm 1.3$
& $833.3 \pm 14.9$ & $0.0861 \pm 0.0028$ \\
\midrule
0.1 & DTS (Null-Reward) & Best-$N$
& $859.0 \pm 30.4$ & $\mathbf{100.0 \pm 0.0}$
& $629.3 \pm 33.9$ & $0.0319 \pm 0.0006$ \\
0.3 & DTS (Null-Reward) & Best-$N$
& $1095.3 \pm 55.4$ & $91.4 \pm 2.2$
& $904.3 \pm 29.4$ & $0.0500 \pm 0.0007$ \\
0.5 & DTS (Null-Reward) & Best-$N$
& $973.7 \pm 26.9$ & $80.0 \pm 1.7$
& $894.7 \pm 27.4$ & $0.0599 \pm 0.0020$ \\
\addlinespace
0.1 & DTS + $\Ehull$ reward & Best-$N$
& $710.3 \pm 25.6$ & $\mathbf{100.0 \pm 0.0}$
& $506.3 \pm 13.1$ & $\mathbf{0.0185 \pm 0.0023}$ \\
0.3 & DTS + $\Ehull$ reward & Best-$N$
& $\mathbf{1207.0 \pm 29.8}$ & $98.3 \pm 1.3$
& $969.7 \pm 27.9$ & $0.0454 \pm 0.0015$ \\
0.5 & DTS + $\Ehull$ reward & Best-$N$
& $1087.3 \pm 22.8$ & $82.4 \pm 1.5$
& $\mathbf{979.7 \pm 15.2}$ & $0.0581 \pm 0.0024$ \\
\bottomrule
\end{tabular}%
}
\end{table}

These results show that given the same NFE but selecting the highest-reward samples, DTS is able to generate far more SUN candidates. Alternatively, for the same target number of SUN crystals, DTS is able to generate using fewer NFEs. However, using a smaller value of $\alpha$ leads to more trajectory reuse and less diversity, which can result in fewer SUN crystals.

\paragraph{DiffCSP.}
We evaluate a matched output count of 1,000 and 10,000 candidates
at the 1M- and 10M-NFE budgets, respectively. We use three protocols
to select these $N$ candidates:
(i) the first $N$ structures generated, which involves no post-hoc
energy selection;
(ii) the $N$ structures with the lowest $\Ehull$; and
(iii) a formula-diverse selection that first maximizes distinct
chemical-formula coverage and then ranks candidates within each
formula by $\Ehull$.

\begin{table}[t]
\centering
\caption{Matched-output comparisons with 1,000 candidates at the
1M-NFE budget and 10,000 candidates at the 10M-NFE budget.}
\label{tab:diffcsp_matched_outputs}
\resizebox{0.8\textwidth}{!}{%
\begin{tabular}{llrrr}
\toprule
Selection & Method & Stability (\%) $\uparrow$
& SUN (\%) $\uparrow$ & Mean $\Ehull$ $\downarrow$ \\
\midrule
\multicolumn{5}{c}{\textbf{Matched 1,000 output candidates}} \\
\midrule
\multirow{4}{*}{First 1,000}
& Best-of-$N$ & $30.7 \pm 0.9$ & $21.0 \pm 1.3$ & $0.314 \pm 0.031$ \\
& DTS (Null-Reward) & $31.3 \pm 0.3$ & $22.2 \pm 1.3$ & $0.349 \pm 0.062$ \\
& \textbf{DTS} & $\mathbf{35.2 \pm 5.6}$ & $\mathbf{24.2 \pm 4.3}$ & $\mathbf{0.278 \pm 0.027}$ \\
& TreeG-SD & $33.5 \pm 4.8$ & $23.2 \pm 3.4$ & $0.317 \pm 0.034$ \\
\addlinespace
\multirow{4}{*}{Lowest $\Ehull$ 1,000}
& Best-of-$N$ & $30.7 \pm 0.9$ & $21.0 \pm 1.3$ & $0.314 \pm 0.031$ \\
& DTS (Null-Reward) & $43.2 \pm 2.0$ & $30.4 \pm 2.8$ & $0.145 \pm 0.002$ \\
& \textbf{DTS} & $\mathbf{49.7 \pm 6.6}$ & $\mathbf{33.9 \pm 4.8}$ & $\mathbf{0.123 \pm 0.018}$ \\
& TreeG-SD & $33.5 \pm 4.8$ & $23.2 \pm 3.4$ & $0.317 \pm 0.034$ \\
\addlinespace
\multirow{4}{*}{Formula-diverse 1,000}
& Best-of-$N$ & $30.7 \pm 0.9$ & $21.0 \pm 1.3$ & $0.314 \pm 0.031$ \\
& DTS (Null-Reward) & $41.9 \pm 1.8$ & $29.9 \pm 2.8$ & $0.151 \pm 0.002$ \\
& \textbf{DTS} & $\mathbf{48.3 \pm 6.4}$ & $\mathbf{33.6 \pm 4.7}$ & $\mathbf{0.128 \pm 0.018}$ \\
& TreeG-SD & $33.5 \pm 4.8$ & $23.2 \pm 3.4$ & $0.317 \pm 0.034$ \\
\midrule
\multicolumn{5}{c}{\textbf{Matched 10,000 output candidates}} \\
\midrule
\multirow{4}{*}{First 10,000}
& Best-of-$N$ & $31.1 \pm 0.3$ & $20.3 \pm 0.1$ & $0.435 \pm 0.035$ \\
& DTS (Null-Reward) & $30.3 \pm 0.6$ & $19.6 \pm 0.9$ & $0.338 \pm 0.020$ \\
& \textbf{DTS} & $\mathbf{40.5 \pm 6.6}$ & $\mathbf{24.4 \pm 2.1}$ & $\mathbf{0.317 \pm 0.125}$ \\
& TreeG-SD & $31.6 \pm 0.5$ & $20.8 \pm 0.0$ & $0.425 \pm 0.033$ \\
\addlinespace
\multirow{4}{*}{Lowest $\Ehull$ 10,000}
& Best-of-$N$ & $31.1 \pm 0.3$ & $20.3 \pm 0.1$ & $0.435 \pm 0.035$ \\
& DTS (Null-Reward) & $47.8 \pm 1.3$ & $30.3 \pm 0.2$ & $0.121 \pm 0.003$ \\
& \textbf{DTS} & $\mathbf{64.2 \pm 10.4}$ & $\mathbf{37.3 \pm 3.5}$ & $\mathbf{0.080 \pm 0.022}$ \\
& TreeG-SD & $31.6 \pm 0.5$ & $20.8 \pm 0.0$ & $0.425 \pm 0.033$ \\
\addlinespace
\multirow{4}{*}{Formula-diverse 10,000}
& Best-of-$N$ & $31.1 \pm 0.3$ & $20.3 \pm 0.1$ & $0.435 \pm 0.035$ \\
& DTS (Null-Reward) & $40.5 \pm 0.9$ & $29.0 \pm 0.3$ & $0.149 \pm 0.002$ \\
& \textbf{DTS} & $\mathbf{50.0 \pm 5.5}$ & $\mathbf{35.7 \pm 3.3}$ & $\mathbf{0.120 \pm 0.014}$ \\
& TreeG-SD & $31.6 \pm 0.5$ & $20.8 \pm 0.0$ & $0.425 \pm 0.033$ \\
\bottomrule
\end{tabular}%
}
\end{table}

SVDD-PM and SMC are omitted only from these matched-output comparisons
because they produce fewer than 1,000 and 10,000 terminal structures
at the corresponding NFE budgets. The first-$N$ protocol shows that DTS's improvement
is not solely due to post-hoc selection. The other protocols
demonstrate the practical benefit of trajectory reuse: at the same
inference budget, DTS produces additional candidates from which a
fixed-size downstream batch can be selected, while the formula-diverse
protocol shows that the improvement persists after controlling for
composition concentration.

\subsection{Dielectric Constant and Bandgap}
More results for the multi-objective dielectric constant vs bandgap experiment are included in \cref{tab:dbg_full}. We compare across different values of progressive widening parameter $\alpha$. The conditional version of MatterGen (DTS-C) is fine-tuned over 200 epochs on the MP-20 dataset, where each crystal is labeled with dielectric constants and bandgaps predicted by ALIGNN \citep{choudhary2021atomistic} and using a conditioning dropout of 20\%.
We use Tchebycheff scalarization throughout this experiment. The conditional model is conditioned on the Dielectric constant and bandgaps of samples from the Pareto front of the training set, determined using Tchebycheff scalarization. Preferences are sampled from a Dirichlet distribution with $\alpha=1$. 
We also compare against a variant that uses an unconditional MatterGen model as its base (DTS-U). 

\begin{table}
\caption{Summary metrics for all evaluated methods on the dielectric/bandgap multi-objective task. $|PF|$ = Pareto front size. For methods with multiple seeds, entries are mean $\pm$ std across seeds. Threshold columns count structures where bandgap\,$\times$\,dielectric $\geq$ threshold.}
\label{tab:dbg_full}
\resizebox{\linewidth}{!}{
\begin{tabular}{lrrrrrrrrrr}
\toprule
 Method & $N$ & $|PF|$ & $N_{\geq 30}$ & $N_{\geq 60}$ & $N_{\geq 120}$ & $N_{\geq 240}$ & HV$\uparrow$ & GD+$\downarrow$ & IGD+$\downarrow$ & R2$\downarrow$ \\
\midrule
Training set & 27,136 & 20 & 4,180 & 1,920 & 526 & 109 & 0.309 & 0.000 & 0.000 & 0.149 \\
MatterGen (cond.) & 665 $\pm$ 0 & 14 $\pm$ 1 & 262 $\pm$ 3 & 138 $\pm$ 6 & 28 $\pm$ 2 & 4 $\pm$ 0 & 0.118 $\pm$ 0.011 & 0.098 $\pm$ 0.024 & 0.143 $\pm$ 0.005 & 0.292 $\pm$ 0.002 \\
DTS-C $\alpha$=0.5 & 645 $\pm$ 40 & 16 $\pm$ 1 & 402 $\pm$ 54 & 234 $\pm$ 27 & 67 $\pm$ 12 & 17 $\pm$ 4 & 0.173 $\pm$ 0.057 & 0.054 $\pm$ 0.015 & 0.180 $\pm$ 0.045 & 0.338 $\pm$ 0.040 \\
DTS-C $\alpha$=0.3 & 674 $\pm$ 44 & 15 $\pm$ 3 & 403 $\pm$ 31 & 231 $\pm$ 23 & 68 $\pm$ 6 & 19 $\pm$ 1 & 0.236 $\pm$ 0.052 & 0.054 $\pm$ 0.012 & 0.120 $\pm$ 0.043 & 0.287 $\pm$ 0.037 \\
DTS-C $\alpha$=0.1 & 804 $\pm$ 46 & 11 $\pm$ 1 & 455 $\pm$ 35 & 242 $\pm$ 49 & 59 $\pm$ 21 & 13 $\pm$ 5 & 0.155 $\pm$ 0.032 & 0.030 $\pm$ 0.025 & 0.214 $\pm$ 0.028 & 0.375 $\pm$ 0.024 \\
DTS-U $\alpha$=0.1 & 1,433 & 12 & 198 & 60 & 9 & 3 & 0.061 & 0.180 & 0.249 & 0.369 \\
DTS-U $\alpha$=0.3 & 1,172 & 16 & 187 & 57 & 6 & 0 & 0.031 & 0.211 & 0.194 & 0.325 \\
DTS-U $\alpha$=0.5 & 1,057 & 23 & 116 & 37 & 2 & 0 & 0.024 & 0.266 & 0.201 & 0.330 \\
\bottomrule
\end{tabular}
}
\end{table}

\subsection{Symmetry-Conditioned Generation}
\label{app:symmetry}
We ran the symmetry-conditioned experiment of \cref{sec:symmetry} with four target space groups, chosen to span a range of rarity in the MatterGen training distribution. All runs use the space-group-conditional MatterGen variant for both DTS and the best-of-$N$ baseline; the reward is the indicator $r(\rvx_0) = \mathbbm{1}[\mathrm{SG}(\rvx_0) = g^\star]$, for target group $g^\star$.
 
\paragraph{Target space groups.}
We selected four space groups that appear rarely in the training set. Their frequencies span  from $\sim$10$^{-2}$ to $\sim$10$^{-5}$, allowing us to see how proposal-distribution coverage interacts with search.
 
\begin{figure}[h]
    \centering
    \includegraphics[width=\linewidth]{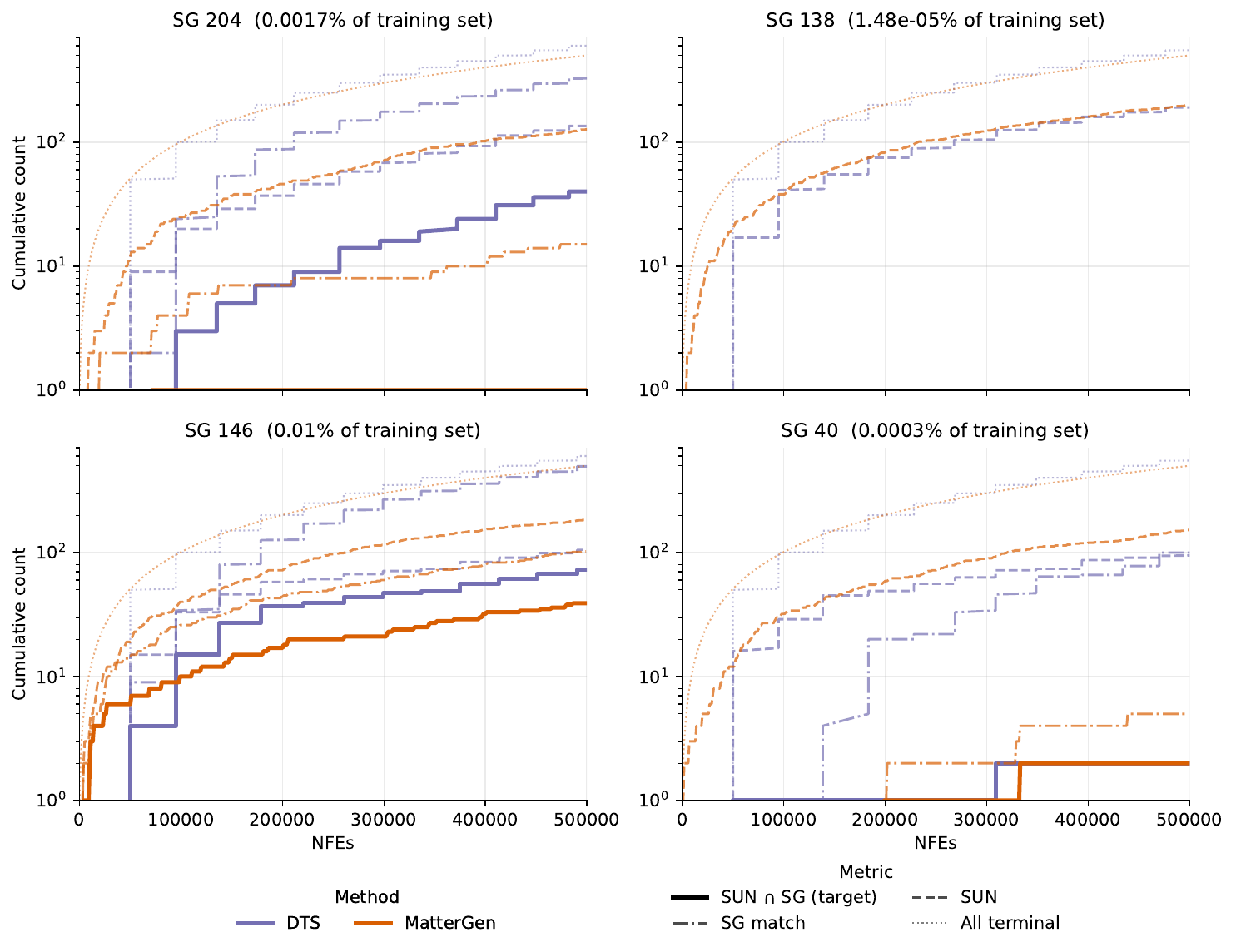}
    \caption{Full view symmetry-conditioned experiment. Color encodes method (DTS vs.\ MatterGen best-of-$N$); linestyle encodes metric.}
    \label{fig:symmetry_appendix}
\end{figure}

\paragraph{Results.}
\cref{fig:symmetry_appendix} shows cumulative SUN $\cap$ target-SG counts versus NFE budget for each target. Three of the four targets (SG 40, 146, 204) follow the pattern of \cref{sec:symmetry}: DTS produces substantially more on-target SUN structures than best-of-$N$, with the gap widening as the budget grows. 
The exception is SG 138, the rarest target at $\sim$10$^{-5}$\,\% of the training set: neither method produces any SUN structures matching this space group within the 500K-NFE budget. 
All runs use $\alpha = 0.5$, $C = 1.0$, $\lambda = 100$. The NFE budget is 500K per method per target. Results are from a single seed per (method, target) pair.

\section{Importance Sampling Details}
\label{sec:is-ablations}
\subsection{Importance Sampling Ablations}
In addition to the ablation study in \Cref{sec:density}, we ran ablations on importance sampling (IS) correction for the Space Group and Dielectric/Bandgap experiments. We note that in our method, the purpose of the correction is not to improve reward or yield. A conditional model can provide a more efficient proposal, but it also changes the proposal distribution. The likelihood-ratio correction is intended to reduce this proposal-induced bias relative to the reward-tilted unconditional target.

\paragraph{Target space group.}
We ablate the correction when generating crystals in space group 204 using a space-group-conditioned MatterGen proposal. We evaluate over 3 seeds and $10^6$ NFEs.

\begin{table}[h]
  \centering
  \caption{IS ablation for space group 204 targeting (3 seeds, 1M NFEs).}
  \label{tab:is-sg204}
  \begin{tabular}{lc}
    \toprule
    Method & \# SG 204 and SUN \\
    \midrule
    DTS with IS    & $88.0 \pm 16.4$ \\
    DTS without IS & $75.0 \pm 24.7$ \\
    \bottomrule
  \end{tabular}
\end{table}

\paragraph{Dielectric constant and bandgap.}
We next ablate IS on the anti-correlated dielectric/bandgap multi-objective task. All runs use $\lambda = 100$ 
and progressive widening $\alpha = 0.5$. 
Note that this evaluation uses only $10^6$ NFEs, and is not directly comparable to the main results of \cref{sec:dielectric}, which used $2.5 \times 10^6$ NFEs.

We compare DTS with importance sampling correction, DTS without IS, and unconditional MatterGen best-of-$N$.

\begin{table}[h]
  \centering
  \caption{IS ablation on the dielectric/bandgap multi-objective task (1M NFEs). $N_{\geq k}$ denotes the number of structures satisfying threshold $k$.}
  \label{tab:is-dielectric}
  \resizebox{\linewidth}{!}{%
  \begin{tabular}{lcccccccc}
    \toprule
    Method & SUN Structures & HV $\uparrow$ & GD+ $\downarrow$ & R2 $\downarrow$ & $N_{\geq 30}$ & $N_{\geq 60}$ & $N_{\geq 120}$ & $N_{\geq 240}$ \\
    \midrule
    Best-of-$N$ & $264 \pm 0$ & $0.072 \pm 0.035$ & $0.183 \pm 0.060$ & $0.306 \pm 0.008$ & $103 \pm 1$ & $57 \pm 3$ & $13 \pm 2$ & $1 \pm 1$ \\
    DTS, no IS            & $325 \pm 34$ & $0.084 \pm 0.021$ & $0.149 \pm 0.022$ & $0.371 \pm 0.024$ & $154 \pm 16$ & $78 \pm 8$ & $20 \pm 4$ & $3 \pm 1$ \\
    DTS, IS               & $286 \pm 12$ & $0.068 \pm 0.024$ & $0.152 \pm 0.037$ & $0.352 \pm 0.013$ & $146 \pm 7$ & $77 \pm 5$ & $20 \pm 3$ & $2 \pm 0$ \\
    \bottomrule
  \end{tabular}%
  }
\end{table}

Removing the IS correction leaves most Pareto metrics within one seed standard deviation of the corrected run, so there is no strong degradation from dropping importance-corrected backup/selection in this setting.

\newcommand{\colhead}[1]{\begin{tabular}{@{}c@{}}#1\end{tabular}}

\subsection{Importance Sampling Statistics}
\label{sec:is-statistics}

We report edge log-weight distributions and normalized effective sample size, $\mathrm{ESS}/n$, for the space-group (\cref{tab:is-stats-sg}), dielectric/bandgap (\cref{tab:is-stats-dielectric}), and density/bulk modulus experiments (\cref{tab:is-stats-density-bulk}). Each tree edge spans 100 diffusion steps between consecutive branching points; its log-weight is the sum of per-step log-ratios over that interval, and the quantiles below are computed over all edges in each interval. At each branching node,
\begin{equation}
  \frac{\mathrm{ESS}}{n} = \frac{\left(\sum_{i=1}^{n} \eta_i\right)^2}{n \sum_{i=1}^{n} \eta_i^2}
\end{equation}
across its $n$ children: a value of 1 indicates equal weights, while a value of approximately $1/n$ indicates concentration on a single child.

\subsubsection{Space group.}

\begin{table}[h]
  \centering
  \caption{Importance sampling statistics for the space group 204 experiment.}
  \label{tab:is-stats-sg}
  \resizebox{\linewidth}{!}{%
  \begin{tabular}{lcccccc}
    \toprule
    Timesteps & \colhead{\# Branching\\nodes} & \colhead{Mean \#\\children} & \colhead{Median\\$\mathrm{ESS}/n$} & \colhead{Log weight\\5\%} & \colhead{Median\\log weight} & \colhead{Log weight\\95\%} \\
    \midrule
    1000--901 & --  & 100.00 & 0.348 & $-0.03$  & $-0.0001$ & 0.03 \\
    900--801  & 169 & 5.29   & 0.763 & $-0.05$  & $-0.0000$ & 0.05 \\
    800--701  & 304 & 5.30   & 0.688 & $-0.08$  & $-0.0001$ & 0.07 \\
    700--601  & 477 & 3.36   & 0.501 & $-0.40$  & $-0.0011$ & 0.36 \\
    600--501  & 188 & 2.88   & 0.500 & $-15.20$ & $-1.40$   & 1.50 \\
    500--401  & 21  & 2.76   & 0.500 & $-19.79$ & $-2.38$   & 0.71 \\
    400--301  & 1   & 4.00   & 0.458 & $-1.65$  & $-0.07$   & 0.58 \\
    \bottomrule
  \end{tabular}%
  }
\end{table}

For most of the trajectory ($t = 1000$--$701$), conditional and unconditional transitions remain closely aligned: the median log-weight is approximately 0, the central 90\% lies roughly within $[-0.08, 0.07]$, and the median $\mathrm{ESS}/n$ is 0.69--0.76 at levels with ${\sim}5$ children. Concentration peaks between $t = 600$ and $401$, where the log-weight tail widens and the median $\mathrm{ESS}/n$ drops to 0.50. This effect is localized rather than trajectory-wide; the $t = 500$--$401$ and $400$--$301$ bins contain only 21 and 1 branching nodes, respectively, so those estimates are noisier.

\subsubsection{Dielectric constant and bandgap.}

\begin{table}[h]
  \centering
  \caption{Importance sampling statistics for the dielectric/bandgap experiment.}
  \label{tab:is-stats-dielectric}
  \resizebox{\linewidth}{!}{%
  \begin{tabular}{lcccccc}
    \toprule
    Timesteps & \colhead{\# Branching\\nodes} & \colhead{Mean \#\\children} & \colhead{Median\\$\mathrm{ESS}/n$} & \colhead{Log weight\\5\%} & \colhead{Median\\log weight} & \colhead{Log weight\\95\%} \\
    \midrule
    1000--901 & 3   & 73.00 & 0.508 & $-0.002$ & $-0.000$ & 0.002 \\
    900--801  & 210 & 6.29  & 0.709 & $-0.002$ & $-0.000$ & 0.002 \\
    800--701  & 481 & 3.59  & 0.795 & $-0.003$ & $-0.000$ & 0.003 \\
    700--601  & 284 & 3.33  & 0.791 & $-0.090$ & $-0.000$ & 0.087 \\
    600--501  & 130 & 3.81  & 0.519 & $-0.681$ & $-0.021$ & 0.482 \\
    500--401  & 76  & 3.78  & 0.547 & $-0.804$ & $-0.028$ & 0.495 \\
    400--301  & 46  & 3.86  & 0.657 & $-0.457$ & $-0.007$ & 0.319 \\
    300--201  & 31  & 3.33  & 0.623 & $-0.370$ & $-0.005$ & 0.273 \\
    200--101  & 15  & 2.98  & 0.717 & $-0.416$ & $-0.006$ & 0.299 \\
    100--001  & 5   & 2.00  & 0.746 & $-0.508$ & $-0.007$ & 0.348 \\
    \bottomrule
  \end{tabular}%
  }
\end{table}

For dielectric/bandgap, the median $\mathrm{ESS}/n$ is 0.71--0.80 between $t = 900$ and $601$, then dips to 0.52--0.55 between $t = 600$ and $401$,
before recovering to 0.62--0.75.

Weights remain usable because DTS corrects locally between consecutive branching nodes rather than accumulating a single importance ratio over the full 1{,}000-step trajectory. Each decision uses only the ratio from the preceding 100-step segment.

Overall, the weights are neither uniformly benign nor fully degenerate: most branching levels retain moderate ESS, and both tasks show a localized mid-trajectory interval with broader log-weight tails and fewer effective children.

\subsubsection{Density/bulk modulus.}

\begin{table}[H]
  \centering
  \caption{Importance sampling statistics for the density/bulk modulus experiment.}
  \label{tab:is-stats-density-bulk}
  \begin{tabular}{lccccc}
    \toprule
    Timesteps & \colhead{\# Branching\\nodes} & \colhead{Median\\$\mathrm{ESS}/n$} & \colhead{Log weight\\5\%} & \colhead{Median\\log weight} & \colhead{Log weight\\95\%} \\
    \midrule
    1000--900 & 48  & 0.740 & $-0.0145$ & $-0.0003$ & 0.0133 \\
    900--800  & 122 & 0.816 & $-0.0145$ & $-0.0001$ & 0.0138 \\
    800--700  & 244 & 0.826 & $-0.0174$ & $-0.0001$ & 0.0167 \\
    700--600  & 314 & 0.658 & $-0.0777$ & $-0.0002$ & 0.0742 \\
    600--500  & 161 & 0.674 & $-0.4631$ & $-0.0093$ & 0.3683 \\
    500--400  & 21  & 0.803 & $-0.4804$ & $-0.0100$ & 0.3521 \\
    400--300  & 2   & 0.729 & $-0.1806$ & $-0.0016$ & 0.1508 \\
    300--200  & 1   & 0.995 & $-0.1350$ & $-0.0013$ & 0.1199 \\
    200--100  & 0   & --    & $-0.1571$ & $-0.0013$ & 0.1384 \\
    100--0    & 0   & --    & $-0.2253$ & $-0.0032$ & 0.1930 \\
    \bottomrule
  \end{tabular}
\end{table}

\end{document}

%% file: math_commands.tex
\usepackage{amsmath,amsfonts,bm}

\def\eqref#1{equation~\ref{#1}}

\def\1{\bm{1}}

\def\rva{{\mathbf{a}}}

\def\rvr{{\mathbf{r}}}

\def\rvv{{\mathbf{v}}}

\def\rvx{{\mathbf{x}}}

\def\rvz{{\mathbf{z}}}

\def\vz{{\bm{z}}}

\DeclareMathAlphabet{\mathsfit}{\encodingdefault}{\sfdefault}{m}{sl}
\SetMathAlphabet{\mathsfit}{bold}{\encodingdefault}{\sfdefault}{bx}{n}

\def\sR{{\mathbb{R}}}

\newcommand{\E}{\mathbb{E}}

